\documentclass[sn-mathphys]{sn-jnl}

\jyear{2021}%

\usepackage{graphicx}%
\usepackage{multirow}%
\usepackage{amsmath,amssymb,amsfonts}%
\usepackage{amsthm}%
\usepackage{mathrsfs}%
\usepackage[title]{appendix}%
\usepackage{xcolor}%
\usepackage{textcomp}%
\usepackage{manyfoot}%
\usepackage{bbding}
\usepackage{cleveref}
\usepackage{booktabs}%
\usepackage{listings}%
\usepackage{ulem}%
\usepackage{longtable}
\usepackage{arydshln}
\usepackage{bm}
\usepackage{comment}
\usepackage{hyperref}
\usepackage{algorithm}
\usepackage[algo2e]{algorithm2e} 
\usepackage{algorithmicx}
\usepackage{algpseudocode}
\usepackage{float}

\usepackage{subcaption}

\theoremstyle{thmstyleone}%
\theoremstyle{thmstyletwo}%

\theoremstyle{thmstylethree}%
\definecolor{darkgreen}{RGB}{0,0,0}

\begin{document}

\title[ ]{Versatile Cardiovascular Signal Generation \\ with a Unified Diffusion Transformer
}


\author[1]{Zehua Chen}\email{zhc23thuml@tsinghua.edu.cn}
\equalcont{These authors contributed equally to this work.}

\author[1,3]{Yuyang Miao}\email{ym520@ic.ac.uk}
\equalcont{These authors contributed equally to this work.}

\author*[1,2]{Liyuan Wang}\email{wly19@tsinghua.org.cn}
\equalcont{These authors contributed equally to this work.}

\author[4]{Luyun Fan}\email{katevan@163.com}

\author[3]{Danilo P. Mandic}\email{d.mandic@imperial.ac.uk}

\author*[1]{Jun Zhu}\email{dcszj@tsinghua.edu.cn}

\affil[1]{Department of Computer Science and Technology, Institute for AI, BNRist Center, THBI Lab, Tsinghua-Bosch Joint Center for ML, Tsinghua University, Beijing, China}

\affil[2]{Department of Psychological and Cognitive Sciences, Tsinghua University, Beijing, China}

\affil[3]{Department of Electrical and Electronic Engineering, Imperial College London, London, United Kingdom}

\affil[4]{Beijing Anzhen Hospital of Capital Medical University, Beijing Institute of Heart Lung and Blood Vessel Diseases, Chinese Institutes for Medical Research, Beijing, China}


\abstract{
Cardiovascular signals such as photoplethysmography (PPG), electrocardiography (ECG), and blood pressure (BP) are inherently correlated and complementary, together reflecting the health of cardiovascular system. However, their joint utilization in real-time monitoring is severely limited by diverse acquisition challenges from noisy wearable recordings to burdened invasive procedures. Here we propose UniCardio, a multi-modal diffusion transformer that reconstructs low-quality signals and synthesizes unrecorded signals in a unified generative framework. Its key innovations include a specialized model architecture to manage the signal modalities involved in generation tasks and a continual learning paradigm to incorporate varying modality combinations. By exploiting the complementary nature of cardiovascular signals, UniCardio clearly outperforms recent task-specific baselines in signal denoising, imputation, and translation. The generated signals match the performance of ground-truth signals in detecting abnormal health conditions and estimating vital signs, even in unseen domains, while ensuring interpretability for human experts. These advantages position UniCardio as a promising avenue for advancing AI-assisted healthcare.
}





\maketitle

\section{Introduction}\label{sec1}

\begin{figure}[H]
    \centering
    \vspace{-0.2cm}
    \includegraphics[width=0.75\linewidth]{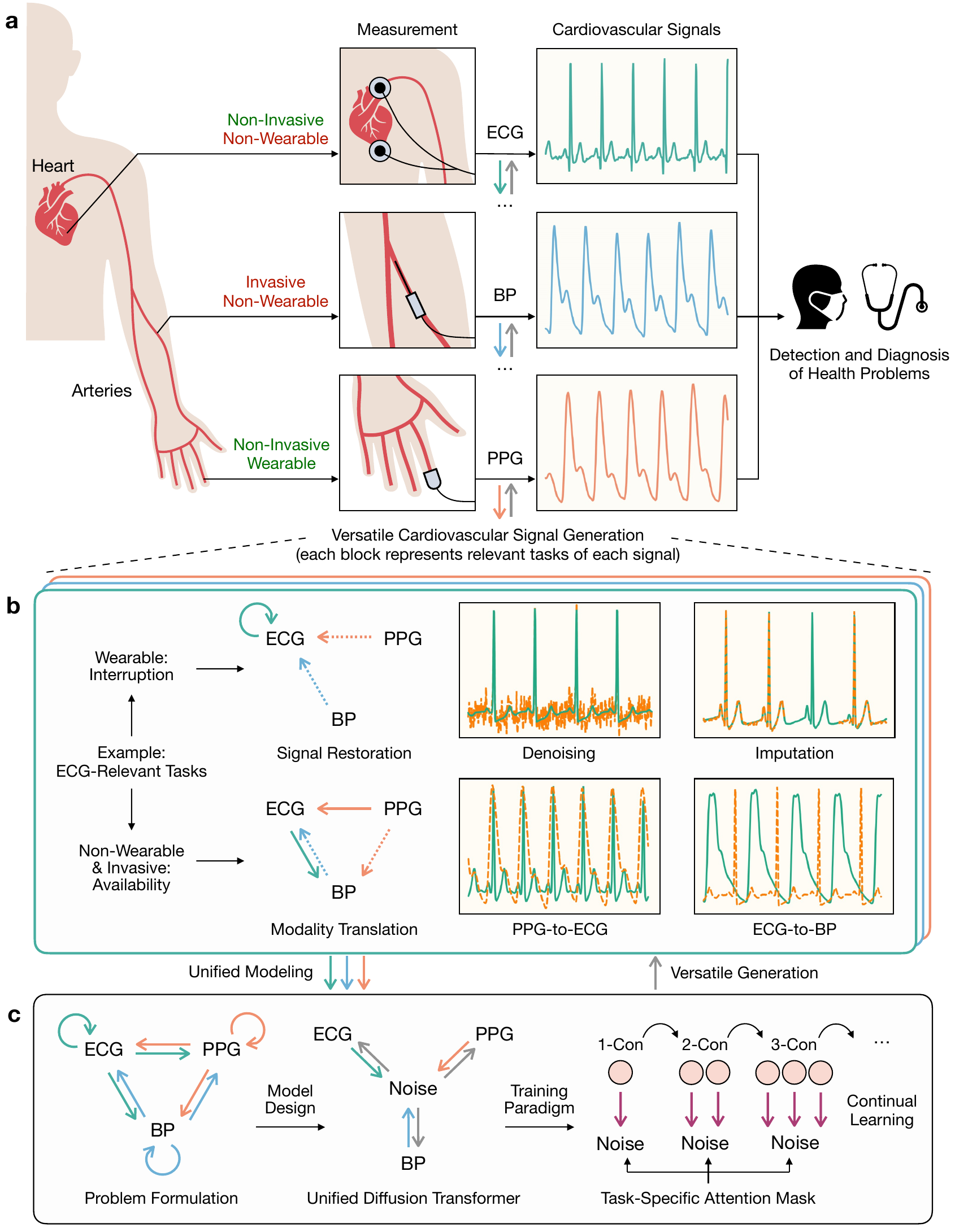}
    \caption{\textbf{Real-time monitoring and diagnosis of cardiovascular signals.} 
    \textbf{a}, Cardiovascular signals, such as PPG, ECG, and BP, are widely used to monitor cardiovascular health and detect abnormalities.
    \textbf{b}, These signals face compromised data quality and availability, which can be addressed by signal restoration and modality translation. Here we take ECG signals as an example, as do the others.
    \textbf{c}, We propose a multi-modal diffusion transformer for versatile cardiovascular signal generation. Our model maps these signals into a unified latent space for flexible use, leveraging task-specific attention masks to regulate modality interactions and a continual learning paradigm to incorporate generation tasks with an increasing number of condition modalities. 
    } 
    \label{Figure1}
    \vspace{-0.6cm}
\end{figure}

Cardiovascular diseases account for nearly 18 million deaths annually, representing 32\% of global mortality~\cite{who2024cvd}. This immense burden underscores the urgent need for effective real-time monitoring to reduce mortality and healthcare costs. Various cardiovascular signals, such as photoplethysmography (PPG)~\cite{alian2014photoplethysmography}, electrocardiography (ECG)~\cite{mirvis2001electrocardiography}, and arterial blood pressure (ABP)~\cite{mcghee2002monitoring}, are commonly used to assess health conditions and detect abnormalities (Fig.~\ref{Figure1}a). PPG signals track blood volume changes within the skin's microvascular tissue, typically recorded at the wrist or fingertips using light-based sensors in wearable devices~\cite{elgendi2024recommendations,tamura2014wearable}. ECG signals monitor the heart's electrical activity by detecting voltage changes in cardiac muscle depolarization and repolarization~\cite{kligfield2007recommendations,trobec2018body}, often requiring precise electrode placement and expert calibration with a sacrifice in wearability. ABP signals are considered the gold standard for blood pressure (BP) assessment, often measured via invasive transducers inserted into arteries~\cite{PASCUAL2013176,saugel2014measurement} with risks of bleeding, infection, and complications.

The diverse challenges of acquiring these signals result in compromised data quality and availability (Fig.~\ref{Figure1}b), which severely limit their joint utilization in healthcare applications. For individuals with relatively normal health conditions, monitoring mainly involves signals obtained from wearable devices, as non-wearable or invasive methods are impractical for routine use. Even in severe health conditions where non-wearable and invasive methods are necessary, prolonged monitoring imposes significant discomfort and patient burdens. Furthermore, the individually recorded signals, particularly those from wearable devices, are susceptible to noise and interruptions, complicating their interpretation by human experts and automated algorithms. Recent efforts have sought to address these challenges by generating cardiovascular signals from recorded ones (Sec.~\ref{sec41}), focusing on individual tasks such as denoising raw recordings~\cite{chiang2019noise,ahmed2023deep}, reconstructing intermittent signals~\cite{xu2022pulseimpute,bansal2021missing}, or translating specific signal pairs (e.g., PPG to ECG)~\cite{shome2024region,sarkar2021cardiogan}. While promising, such task-specific designs fail to exploit the complementary information inherent across distinct signals, resulting in limited efficacy and applicability. 

Given the highly correlated physiological activities of the cardiovascular system, cardiovascular signals correspond to different modalities of a shared underlying process. Modeling their multi-modal conditional distributions can capture the relationships between available recorded signals (condition modalities) and desired generated signals (target modalities), thereby unifying potential generation tasks involving arbitrary signal modalities.
In this work, we propose UniCardio, a specialized diffusion transformer designed to leverage multi-modal relationships among cardiovascular signals, ensuring effective and versatile signal generation (Fig.~\ref{Figure1}c). UniCardio adopts a transformer-based architecture with modality-specific encoders and decoders, alongside task-specific attention masks to regulate modality interactions. Different from other deep generative models~\cite{goodfellow2020generative,de2021diffusion}, UniCardio captures multi-modal conditional distributions with a unified noise-to-data generative framework, mapping intra- and inter-modal relationships into a unified latent space for flexible use. 
To cope with varying combinations of condition and target modalities, UniCardio introduces a continual learning paradigm to incorporate generation tasks involving an increasing number of condition modalities, thereby accommodating progressively complex relationships while explicitly balancing their contributions.

We pre-train UniCardio on the Cuffless BP dataset~\cite{kachuee2016cuffless}, which comprises 339 hours of trimodal recordings of PPG, ECG, and BP signals collected from ICU patients with diverse abnormal health conditions. 
Through joint utilization of multiple cardiovascular signals, UniCardio demonstrates state-of-the-art performance across a range of generation tasks, including denoising, imputation, and translation, outperforming recent task-specific baselines. The generated signals have promoted downstream cardiovascular applications over multiple unseen domains, including detection of abnormal health conditions and estimation of vital signs, achieving comparable or even better performance than the ground-truth signals. The generated signals further ensure interpretability through displaying diagnostic characteristics of typical abnormalities, validated by clinician assessments.
To our knowledge, UniCardio represents the first unified framework for cardiovascular signal generation. 
\textcolor{darkgreen}{It integrates technical innovations in multi-modal generation and continual learning into practical solutions for signal restoration and modality translation, serving as a versatile foundation for advancing real-time health monitoring.}

\section{Results}\label{sec2}
\subsection{Unified multi-modal generative modeling for cardiovascular signals}

To capitalize on the interrelated nature of cardiovascular signals for versatile generation, UniCardio fits their multi-modal conditional distributions to model many-to-any relationships between condition modalities and target modality, thereby unifying signal restoration and modality translation within a shared framework (Fig.~\ref{Figure1}b, Sec.~\ref{sec41}). We obtain this objective with an advanced design of conditional diffusion models~\cite{DDPM2020,SGMs2021}, which offer key advantages for generative modeling. These models operate through a forward process that gradually adds noise to the data, transforming it into a simple prior distribution (typically Gaussian), and a reverse process that learns to reconstruct the data by iteratively removing the noise. UniCardio develops an unconditional forward process that transforms different signal modalities into a unified prior, and a conditional reverse process that guides the iterative reconstruction of desired signals using conditional information alongside a unified diffusion step, enabling flexible handling of diverse condition modalities in generation tasks (Fig.~\ref{Figure1}c, Sec.~\ref{sec42}). The conditional distributions are learned with transformer-based architectures~\cite{bao2023one,peebles2023scalable,UViT2023} to accommodate various input-output configurations (Sec.~\ref{sec43}, Supplementary Fig.~\ref{fig:model_architecture}), with key innovations including modality-specific encoders of multi-scale convolution, customized transformer modules with task-specific attention masks, and modality-specific decoders. 

\begin{figure}[t]
    \centering
    \vspace{-0.1cm}
    \includegraphics[width=0.98\linewidth]{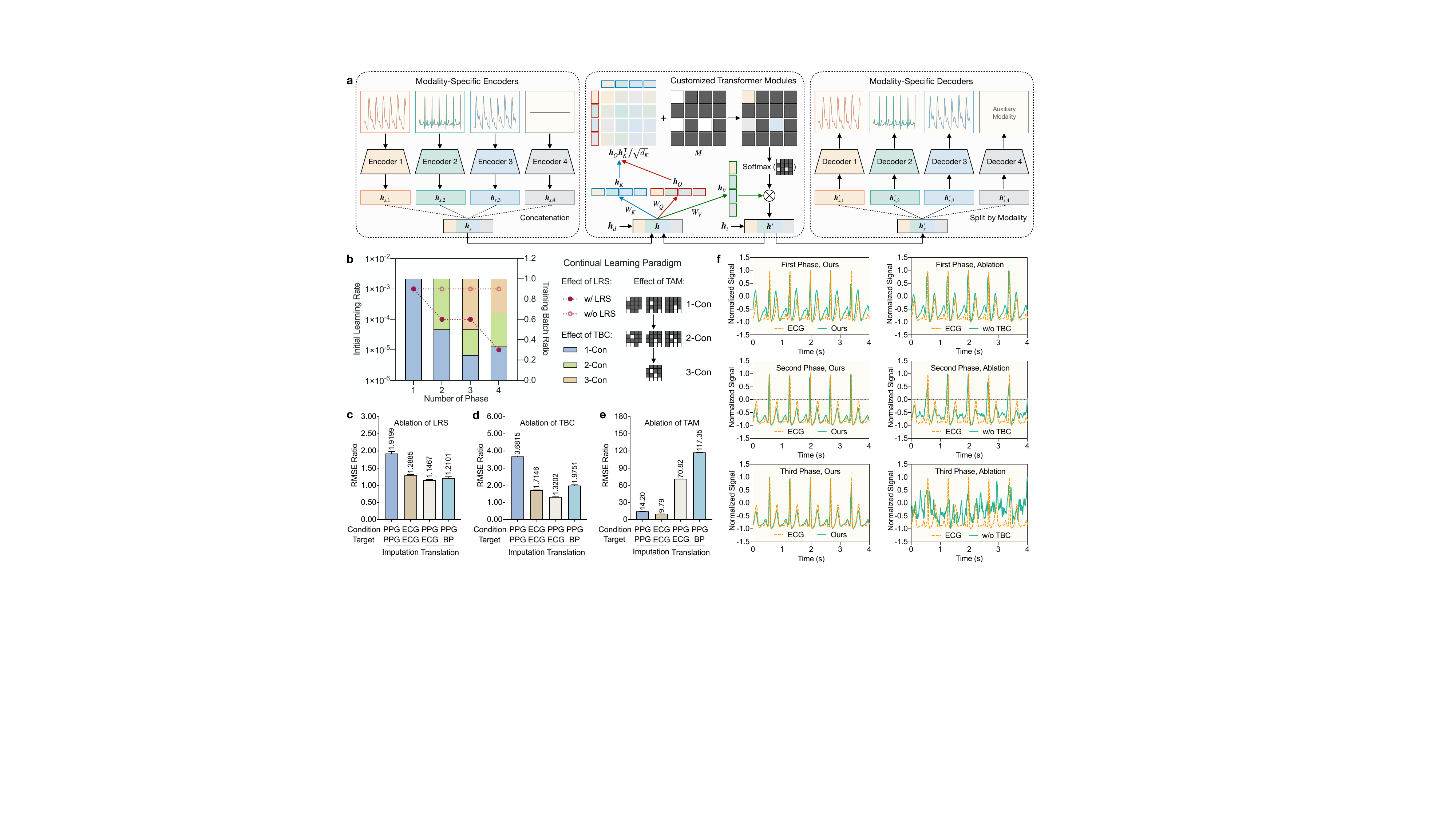}
    \caption{\textbf{Model architecture and training paradigm.}
    \textbf{a}, UniCardio comprises modality-specific encoders, customized transformer modules with task-specific attention masks, and modality-specific decoders. 
    $\bm{h}_s$, concatenated signal representations. $\bm{h}_s'$, generated signal representations. $\bm{h}$, module-wise inputs. $\bm{h}'$, module-wise outputs.
    $\bm{h}_d$, diffusion embeddings. $\bm{h}_t$, time embeddings. 
    The key $\bm{h}_K$, query $\bm{h}_Q$, and value $\bm{h}_V$ are obtained from $\bm{h}$ with the corresponding learnable matrices $W_K$, $W_Q$, and $W_V$.
    \textbf{b}, The continual learning paradigm of generation tasks with an increasing number of condition modalities (1-Con, 2-Con, 3-Con, etc.), achieved by learning rate scheduling (LRS), training batch composition (TBC), and task-specific attention masks (TAM). 
    \textbf{c}-\textbf{e}, Ablation studies of LRS, TBC, and TAM, respectively. 
    The evaluation metric is the RMSE ratio after and before ablation, where the values above 1 indicate worse performance.
    The quantification results are averaged by 256 independent trials. The error bars represent the standard error of the mean.
    \textbf{f}, Phase-wise visualization of catastrophic forgetting, using PPG-to-ECG translation under TBC ablation as an example. 
    } 
    \label{Figure2}
    \vspace{-0.6cm}
\end{figure}

As shown in Fig.~\ref{Figure2}a, the input signals for each modality are processed by the modality-specific encoders, which consist of multiple one-dimensional (1D) convolutional neural networks (CNNs) with various kernel sizes to extract representations across different time scales. The extracted representations of all modalities are concatenated and fed into the customized transformer modules, which further receive learnable diffusion embeddings to encode the current diffusion step and time embeddings to encode the timestamp of each signal point. Through self-attention mechanisms, the transformer modules enable signal points from different timestamps and modalities to share relevant information. To control the modalities involved in specific generation tasks, we implement task-specific attention masks that block the information flow of irrelevant tokens other than the condition-to-target ones. The output representations of multiple customized transformer modules are combined in a residual manner, split by modality, and then projected into 1D generated signals via modality-specific decoders, which are implemented as standard multi-layer perceptrons (MLPs). 

Given multiple cardiovascular signals as the modalities of interest, arbitrary combinations of condition and target modalities can create massive generation tasks with complex relationships, making it highly nontrivial to assign sufficient training samples for each task and properly balance the task-specific loss (Sec.~\ref{sec41}). 
To address this, we propose a continual learning paradigm that incorporates generation tasks with an increasing number of condition modalities in multiple phases, ensuring sufficient training samples and balanced loss weights in each phase. We employ a combination of simple yet effective strategies (Fig.~\ref{Figure2}b, Sec.~\ref{sec44}) to overcome catastrophic forgetting, a well-known issue in continual learning~\cite{wang2024comprehensive,parisi2019continual}. 
These strategies include (1) learning rate scheduling, where the learning rate starts high for effective initialization and decreases gradually in later phases to stabilize knowledge; (2) training batch composition, where the training batches of the current phase are supplemented with a portion of tasks from previous phases to reinforce earlier learning; and (3) task-specific attention masks, which naturally support continual learning by preventing inter-task interference.

Ablation studies validate the efficacy of these strategies (Fig.~\ref{Figure2}c-f, Supplementary Table~\ref{tab:forgetting}). Removing either the learning rate scheduling (Fig.~\ref{Figure2}c), training batch composition (Fig.~\ref{Figure2}d), or task-specific attention masks (Fig.~\ref{Figure2}e) results in a significant increase in the root mean squared error (RMSE) across a variety of generation tasks. In particular, removing task-specific attention masks destroys almost the entire model performance, highlighting their critical role in capturing conditional distributions.
Detailed visualization examples further illustrate the effect of alleviating catastrophic forgetting (Fig.~\ref{Figure2}f). The generated signals with full strategies become progressively aligned with ground-truth signals as the model learns from additional modalities during continual learning. In contrast, the generated signals with an ablation become progressively less informative as more phases are introduced. Interestingly, these strategies correspond to common continual learning methods in terms of regularization~\cite{kirkpatrick2017overcoming,wang2021afec}, replay~\cite{rebuffi2017icarl,wang2021memory}, and architecture~\cite{serra2018overcoming,wang2023incorporating}, suggesting a natural fit between UniCardio's model design and training paradigm.

\begin{figure}[t]
    \centering
    \vspace{-0.1cm}
    \includegraphics[width=0.98\linewidth]{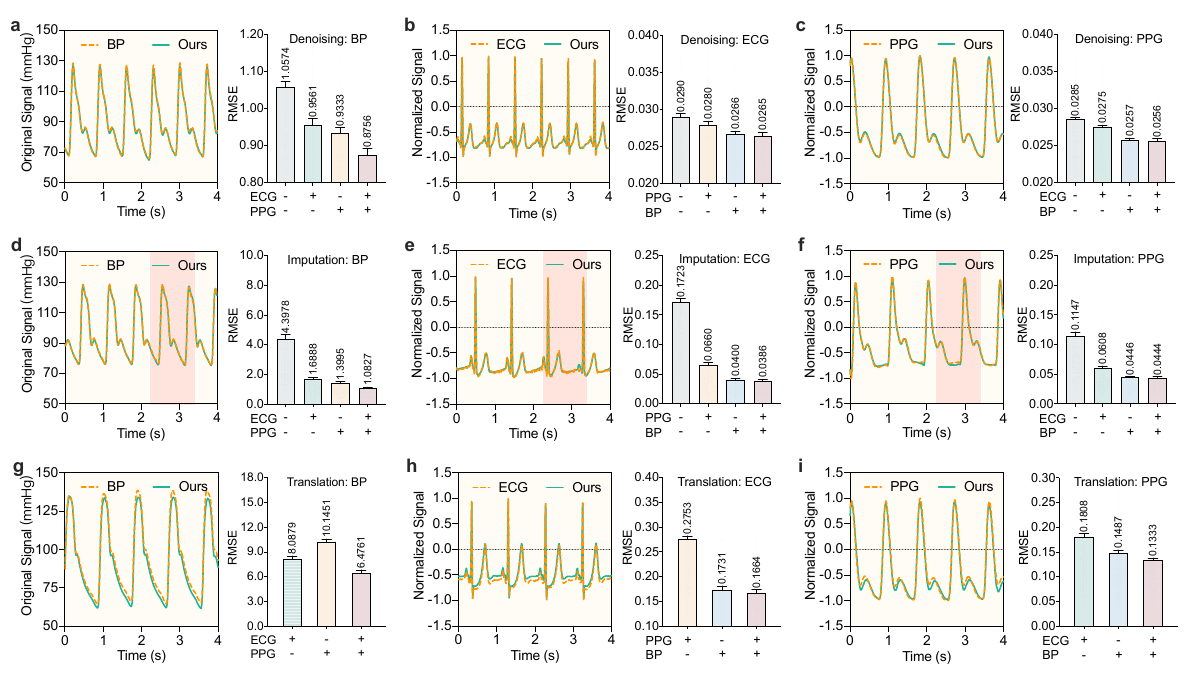}
    \caption{\textbf{Overall performance of versatile generation tasks.} 
    \textbf{a}-\textbf{c}, Denoising tasks, restoring clean signals from noisy raw recordings in each modality.
    \textbf{d}-\textbf{f}, Imputation tasks, reconstructing missing segments (i.e., the masked regions) from intermittent signals.
    \textbf{g}-\textbf{i}, Translation tasks, synthesizing signals of a target modality from one or more condition modalities. 
    For visualization, we present the original values for BP while the normalized values for PPG and ECG.
    The quantification results are averaged by 256 independent trials. The error bars represent the standard error of the mean.
    } 
    \label{Figure3}
    \vspace{-0.5cm}
\end{figure}

\newcommand{\tabincell}[2]{\begin{tabular}{@{}#1@{}}#2\end{tabular}}
\begin{table*}[t]
    \centering
    \caption{\textbf{Comparison of UniCardio against task-specific baselines.} 
    We benchmark UniCardio on four well-studied yet challenging tasks, including PPG imputation, ECG imputation, PPG-to-ECG translation, and PPG-to-BP translation. We report the performance of UniCardio after the multi-modal pre-training, alongside two enhanced variants including UniCardio-F for further fine-tuning on the task of interest and UniCardio-M for incorporating more condition modalities during the testing stage. 
    \textcolor{darkgreen}{Performance metrics include RMSE, Min-RMSE, MAE, Min-MAE, and KS-Test (lower values indicate better performance), averaged by 256 independent trials.} The error bars represent the standard error of the mean.
    ``-I'' denotes the intermittent signals. 
    \textcolor{darkgreen}{The best performance for each task is highlighted in bold.}
    }
    \renewcommand\arraystretch{1.4}
    \resizebox{0.98\textwidth}{!}{ 
    \begin{tabular}{c|l|c|c|c|c|c|c|c}
    \hline
    Task & \,\,\,\,\,\,\, Method & Input Signal & RMSE ($\downarrow$) & Min-RMSE ($\downarrow$) & MAE ($\downarrow$) & Min-MAE ($\downarrow$) & KS-Test ($\downarrow$) & Model Size ($\downarrow$) \\
    \hline
    \multirow{5}*{\tabincell{c}{PPG \\ Imputation}} 
        & PulseImpute~\cite{xu2022pulseimpute} & PPG-I & 0.0763 $\pm$ 0.0036 & 0.0393 $\pm$ 0.0018 & 0.0571 $\pm$ 0.0023 & 0.0164 $\pm$ 0.0007 & 0.1085 $\pm$ 0.0034 & 3.06M  \\ 
        & DeepMVI~\cite{bansal2021missing} & PPG-I & 0.1745 $\pm$ 0.0063 & 0.0890 $\pm$ 0.0030 & 0.1268 $\pm$ 0.0044 & 0.0382 $\pm$ 0.0013 & 0.1395 $\pm$ 0.0042 & \textbf{2.78}M  \\ 
        \cdashline{2-9}[2pt/2pt]
        & UniCardio &  PPG-I & 0.1146 $\pm$ 0.0063 & 0.0585 $\pm$ 0.0031 & 0.0797 $\pm$ 0.0043 & 0.0244 $\pm$ 0.0013 & 0.1084 $\pm$ 0.0039 & 5.51M \\
        & UniCardio-F &  PPG-I & 0.0710 $\pm$ 0.0043 & \textbf{0.0226} $\pm$ 0.0005 & 0.0515 $\pm$ 0.0028 & \textbf{0.0104} $\pm$ 0.0002 & 0.0918 $\pm$ 0.0030 & 5.51M \\
        & UniCardio-M & PPG-I, ECG, BP & \textbf{0.0443} $\pm$ 0.0016 & 0.0243 $\pm$ 0.0009 & \textbf{0.0347} $\pm$ 0.0012 & \textbf{0.0104} $\pm$ 0.0004 & \textbf{0.0907} $\pm$ 0.0031 & 6.09M \\
        \hline
    \multirow{5}*{\tabincell{c}{ECG \\ Imputation}} 
        & PulseImpute~\cite{xu2022pulseimpute} & ECG-I & 0.1391 $\pm$ 0.0033 & 0.0760 $\pm$ 0.0018 & 0.1224 $\pm$ 0.0027 & 0.0367 $\pm$ 0.0008 & 0.4679 $\pm$ 0.0089 & 3.06M \\ 
        & DeepMVI~\cite{bansal2021missing} & ECG-I & 0.2420 $\pm$ 0.0031 & 0.1267 $\pm$ 0.0017 & 0.1395 $\pm$ 0.0022 & 0.0413 $\pm$ 0.0007 & 0.2400 $\pm$ 0.0061 & \textbf{2.78}M \\ 
        \cdashline{2-9}[2pt/2pt]
        & UniCardio & ECG-I & 0.1756 $\pm$ 0.0042 & 0.0891 $\pm$ 0.0025 & 0.0750 $\pm$ 0.0026 & 0.0251 $\pm$ 0.0009 & 0.1486 $\pm$ 0.0055 & 5.51M \\
        & UniCardio-F & ECG-I & 0.0938 $\pm$ 0.0045 & 0.0622 $\pm$ 0.0022 & 0.0448 $\pm$ 0.0024 & 0.0177 $\pm$ 0.0006 & 0.1199 $\pm$ 0.0037 & 5.51M \\
        & UniCardio-M & PPG, ECG-I, BP & \textbf{0.0385} $\pm$ 0.0027 & \textbf{0.0210} $\pm$ 0.0015 & \textbf{0.0241} $\pm$ 0.0017 & \textbf{0.0073} $\pm$ 0.0005 & \textbf{0.1024} $\pm$ 0.0032 & 6.09M \\
        \hline
    \multirow{5}*{\tabincell{c}{PPG-to-ECG \\ Translation}} 
        & RDDM~\cite{shome2024region} & PPG & 0.5710 $\pm$ 0.0140 & 0.5233 $\pm$ 0.0148 & 0.5155 $\pm$ 0.0153 & 0.4839 $\pm$ 0.0153& 0.7706 $\pm$ 0.0137 & 138.77M  \\ 
        & CardioGAN~\cite{sarkar2021cardiogan} & PPG & 0.4313 $\pm$ 0.0100 & 0.3105 $\pm$ 0.0103 & 0.3226 $\pm$ 0.0104 & 0.2598 $\pm$ 0.0103 & 0.5208 $\pm$ 0.0146 & 5.97M  \\ 
        \cdashline{2-9}[2pt/2pt]
        & UniCardio & PPG & 0.2747 $\pm$ 0.0067 & 0.2173 $\pm$ 0.0069 & 0.1937 $\pm$ 0.0070 & 0.1668 $\pm$ 0.0070 & 0.4407 $\pm$ 0.0154 & \textbf{5.51}M \\
        & UniCardio-F & PPG & 0.1960 $\pm$ 0.0062 & \textbf{0.1326} $\pm$ 0.0056 & \textbf{0.1165} $\pm$ 0.0059 & \textbf{0.0888} $\pm$ 0.0056 & \textbf{0.2698} $\pm$ 0.0131 &  \textbf{5.51}M \\
        & UniCardio-M & PPG, BP & \textbf{0.1663} $\pm$ 0.0076 & 0.1396 $\pm$ 0.0071 & 0.1199 $\pm$ 0.0070 & 0.1091 $\pm$ 0.0068 & 0.3302 $\pm$ 0.0147 & 6.09M \\
        \hline
    \multirow{5}*{\tabincell{c}{PPG-to-BP \\ Translation}} 
        & ABPNet~\cite{paviglianiti2020noninvasive} & PPG & 7.2994 $\pm$ 0.2725 & 6.1959 $\pm$ 0.2566 & 5.6647 $\pm$ 0.2434 & 4.8927 $\pm$ 0.2336 & 0.1745 $\pm$ 0.0071  & \textbf{1.59}M \\ 
        & PPG2ABP~\cite{ibtehaz2022ppg2abp} & PPG & 6.1882 $\pm$ 0.2584 & 4.9102 $\pm$ 0.2419 & 4.7835 $\pm$ 0.2076 & \textbf{3.4902} $\pm$ 0.2000 & 0.1621 $\pm$ 0.0060 & 19.44M \\ 
        \cdashline{2-9}[2pt/2pt]
        & UniCardio & PPG & 10.1538 $\pm$ 0.4213 & 8.8916 $\pm$ 0.3937 & 8.3721 $\pm$ 0.3889 & 7.5386 $\pm$ 0.3690 & 0.2668 $\pm$ 0.0097 & 5.51M \\ 
        & UniCardio-F & PPG & \textbf{5.7924} $\pm$ 0.3250 & \textbf{4.5986} $\pm$ 0.2993 & \textbf{4.5893} $\pm$ 0.2897 & 3.8438 $\pm$ 0.2755 & \textbf{0.1567} $\pm$ 0.0079 & 5.51M \\ 
        & UniCardio-M & PPG, ECG & 6.5066 $\pm$ 0.3443 & 5.9460 $\pm$ 0.3206 & 5.4908 $\pm$ 0.3077 & 5.0413 $\pm$ 0.2923 & 0.2004 $\pm$ 0.0086 & 6.09M \\
        \hline
    \end{tabular}
    }
     \vspace{-0.3cm}
\label{Table1}
\end{table*}

\subsection{Versatile high-quality cardiovascular signal generation}

To demonstrate the advantages of unifying multiple signal modalities within a shared framework, we first evaluate UniCardio on three representative generation tasks of practical significance, including denoising, imputation, and translation. These tasks involve diverse input-output configurations and simulate varying degrees of signal degradation, reflecting common challenges in healthcare applications.
Denoising tasks restore clean signals from their noisy raw recordings, which may be affected by powerline interference and muscle contractions~\cite{bing2024novel}. Imputation tasks reconstruct missing segments from intermittent signals, such as filling gaps caused by temporary sensor disconnections or interruptions during long-term monitoring~\cite{xu2022pulseimpute}. 
Translation tasks synthesize signals of a target modality from one or more condition modalities, enables non-invasive or wearable alternatives to traditionally invasive or non-wearable measurements~\cite{shome2024region,sarkar2021cardiogan}.
From denoising, imputation to translation, the generation tasks become progressively more challenging as the degree of signal degradation increases, requiring greater reliance on complementary information from other modalities.

In Fig.~\ref{Figure3}, we employ PPG, ECG, and BP as the target modality during the testing stage, respectively. We present visualization results of the generated signals alongside the ground-truth signals, as well as quantitative results to evaluate the average difference between them. Denoising and imputation tasks (Fig.~\ref{Figure3}a-c and d-f) involve partial signal degradation, with the target modality also included as a condition modality. The generated signals closely match the ground-truth signals, and achieve strong performance with only one condition modality (i.e., the degraded version of target modality). Incorporating additional condition modalities further reduces the average differences, underscoring the benefit of multi-modal relationships. Translation tasks (Fig.~\ref{Figure3}g-i) involve complete signal degradation and are inherently more challenging. These tasks rely entirely on additional condition modalities to generate the target modality. Despite the difficulty, UniCardio produces high-quality signals with only minor deviations from the ground-truth signals. Again, incorporating more condition modalities further improves the performance. 

\textcolor{darkgreen}{
To further investigate the benefit of joint modeling, we conduct an ablation study: Baseline-1, single-modality models trained on only one-condition imputation tasks; Baseline-2, dual-modality models trained on one-condition translation tasks and two-condition imputation tasks; and Baseline-3, tri-modality models trained on two-condition translation tasks and three-condition imputation tasks. All models were trained for the same number of effective epochs to ensure fairness. As shown in Supplementary Table~\ref{tab:joint_model}, UniCardio achieves comparable or slightly inferior performance to Baseline-1 on one-condition imputation and Baseline-2 on one-condition translation, despite being a compact model that handles a much broader set of tasks across all modality combinations. Notably, UniCardio significantly outperforms Baseline-2 on two-condition imputation tasks (31.69\% RMSE ratio), and outperforms Baseline-3 on two-condition translation tasks (35.97\% RMSE ratio) and three-condition imputation tasks (63.85\% RMSE ratio). These results highlight knowledge transfer across task-modality combinations: training on tasks involving fewer modalities facilitates tasks involving more modalities.
}

While UniCardio supports versatile generation tasks across all modalities of pre-training data, we benchmark its performance on four particularly challenging tasks well-studied in the literature, including PPG imputation, ECG imputation, PPG-to-ECG translation, and PPG-to-BP translation. These tasks vary in difficulty, with more complex tasks (e.g., translation compared to imputation) demanding stronger task-specific customizations. This trend is reflected in recent strong baselines. For example, PulseImpute~\cite{xu2022pulseimpute} and DeepMVI~\cite{bansal2021missing} are designed for PPG/ECG imputation, RDDM~\cite{shome2024region} and CardioGAN~\cite{sarkar2021cardiogan} are designed for PPG-to-ECG translation, and ABPNet~\cite{paviglianiti2020noninvasive} and PPG2ABP~\cite{ibtehaz2022ppg2abp} are designed for PPG-to-BP translation.

As shown in Table~\ref{Table1}, UniCardio after multi-modal pre-training has already achieved competing or even better performance across all tasks compared to the corresponding baselines. 
We consider UniCardio's two variants to explore additional enhancements: The first variant UniCardio-F fine-tunes the model using previous pre-training data for a specific generation task, making it more tailored to the task of interest. The second variant UniCardio-M employs more available condition modalities during the testing stage to leverage their complementary information. Both variants deliver substantial improvements, largely outperforming all baselines in most cases. \textcolor{darkgreen}{
To provide additional reference, we also evaluate classical signal processing baselines, such as a wavelet denoising algorithm~\cite{sendur2003bivariate} for denoising and a Gerchberg–Papoulis algorithm~\cite{gerchberg1974super,papoulis2003new} for imputation (Supplementary Table~\ref{tab:classical_methods}). 
As expected, these traditional methods clearly underperform UniCardio as well as other deep learning–based methods.}

Moreover, UniCardio performs all tasks with a comparably small amount of parameters, where the use of different modalities only requires adding the corresponding encoders and decoders (around 0.3M for each modality), making it applicable to wearable devices. In contrast, implementing these functions with task-specific methods would require combining multiple proprietary models, leading to tens of times greater parameter overhead. Such efficiency underscores the particular benefits of incorporating multi-modal relationships within a unified framework, as UniCardio does.

\subsection{Robust real-time health monitoring with UniCardio}

UniCardio's integrated functions of signal restoration and modality translation provide a comprehensive enhancement to cardiovascular signal processing (Fig.~\ref{Figure1}). To demonstrate its practical effectiveness, we apply UniCardio to the publicly available datasets of unseen domains and explore representative applications spanning two major areas: detecting abnormal health conditions and estimating vital signs. Unless otherwise specified, we implement the generated signals without additional fine-tuning. Depending on the specific characteristics of each dataset (e.g., availability of multi-modal signals or disease annotations) as well as the unique demands of each application, we assess UniCardio's generative capabilities in a targeted manner, centering around ECG as a typical example in cardiovascular diagnostics. 

\begin{figure}[t]
    \centering
    \vspace{-0.1cm}
    \includegraphics[width=0.85\linewidth]{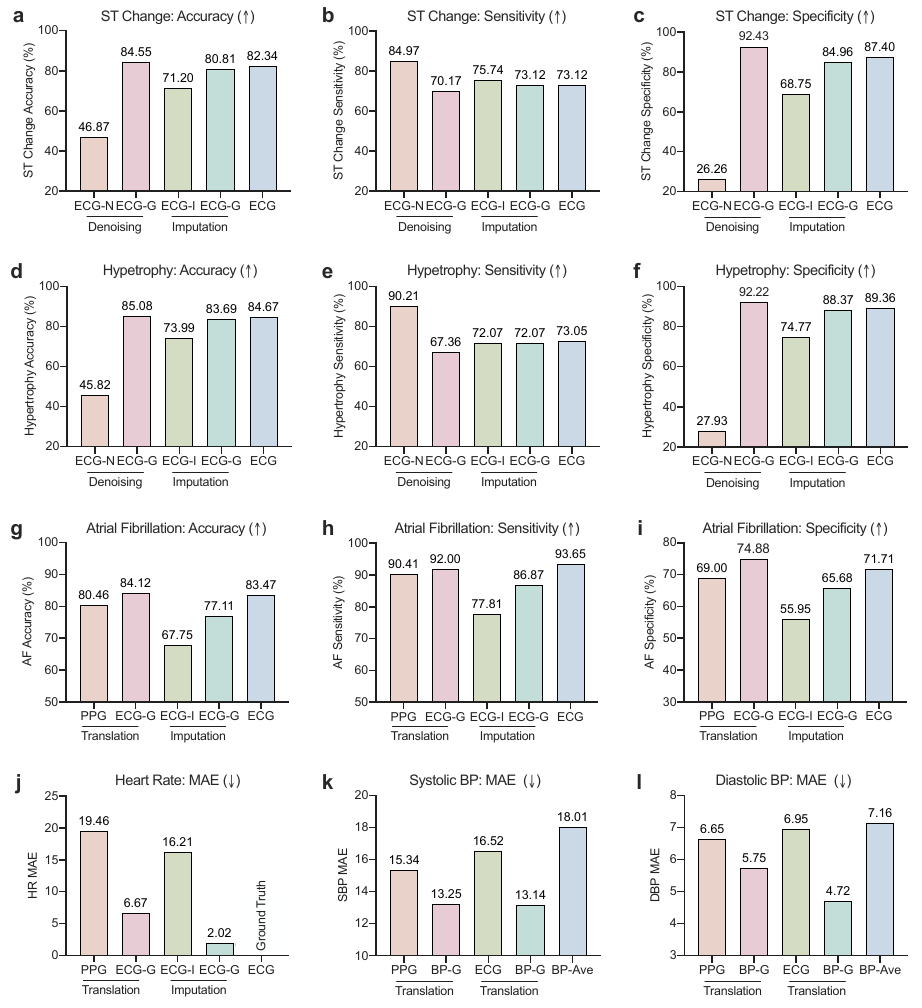}
    \caption{\textbf{Versatile generation assisted cardiovascular applications.}
    \textbf{a}-\textbf{c}, Detection of ST change using the PTBXL dataset~\cite{wagner2020ptb}.
    \textbf{d}-\textbf{f}, Detection of hypertrophy using the PTBXL dataset~\cite{wagner2020ptb}.
    \textbf{g}-\textbf{i}, Detection of atrial fibrillation (AF) using the MIMIC PERform AF dataset~\cite{charlton2022detecting}.
    \textbf{j}, Estimation of heart rate (HR) using the WESAD dataset~\cite{schmidt2018introducing}. 
    \textbf{k}-\textbf{l}, Estimation of systolic BP (SBP) and diastolic BP (SBP) \textcolor{darkgreen}{using the MIMIC dataset~\cite{moody1996database}.} 
    ``-N'' indicates noisy signals. ``-I'' indicates intermittent signals. ``-G'' indicates generated signals. \textcolor{darkgreen}{``-Ave'' indicates using the average SBP/DBP of the training set as the prediction.}
    Here we do not include the error bar since most cases are averaged over the entire test set without repeated sampling.
    } 
    \label{Figure4}
    \vspace{-0.7cm}
\end{figure}

We first evaluate the detection of multiple cardiovascular abnormalities using the PTBXL dataset~\cite{wagner2020ptb}, which includes only ECG signals annotated with abnormal health conditions. Of these, we adopt the ST change (Fig.~\ref{Figure4}a-c) and hypertrophy (Fig.~\ref{Figure4}d-f) as representative examples. This scenario focuses on UniCardio's denoising capability, which mitigates noise-induced degradation during recording and transmission. The noisy signals exhibit low accuracy, extremely low specificity, and inflated sensitivity due to excessive false-positive classifications. In contrast, the denoised signals generated by UniCardio resolve these issues, achieving performance comparable to the ground-truth signals. Additionally, missing segments that may result from sensor disconnections can moderately affect both abnormalities, which are addressed by UniCardio's imputation capability.

We further assess the detection of atrial fibrillation (AF) using the MIMIC PERform AF dataset~\cite{charlton2022detecting}, which includes both PPG and ECG signals annotated with AF (Fig.~\ref{Figure4}g-i). AF is a common cardiac arrhythmia associated with an elevated risk of stroke and heart failure, demanding accurate and timely detection. UniCardio addresses two parallel challenges in this scenario: translating high-quality ECG signals from wearable PPG signals and imputing intermittent ECG signals to restore missing segments. The generated ECG signals significantly outperform both PPG and intermittent ECG signals in terms of accuracy, sensitivity, and specificity. 

\textcolor{darkgreen}{
Next, we evaluate the estimation of vital signs, focusing on heart rate (HR) using the WESAD dataset~\cite{schmidt2018introducing} and systolic/diastolic BP (SBP/DBP) using the MIMIC dataset~\cite{moody1996database}.
These multi-modal datasets allow us to examine the benefits of obtaining non-wearable signals from wearable devices. 
For HR estimation, UniCardio can effectively translate high-quality ECG signals from wearable PPG signals and impute intermittent ECG signals to restore missing segments, substantially reducing the mean absolute error (MAE) compared to the original PPG and degraded ECG signals, respectively (Fig.~\ref{Figure4}j).
For BP estimation, we focus on translation tasks, as intermittent signals have limited impact on SBP/DBP prediction.
Although UniCardio pretrained on Cuffless BP can be directly applited to MIMIC and generate high-quality waveforms (Supplementary Fig.~\ref{BP_Trans_Dataset}), we find that fine-tuning is necessary to obtain accurate SBP/DBP values. The fine-tuned UniCardio achieves consistent improvements over baseline models that predict SBP/DBP directly from PPG and ECG on a strictly held-out MIMIC test set with no subject overlap (Fig.~\ref{Figure4}k,l). We also consider a naive baseline that simply uses the average SBP/DBP from the training set for prediction, which likewise underperforms ours (Fig.~\ref{Figure4}k,l).
}

\textcolor{darkgreen}{
Interestingly, we observe that UniCardio's ECG-to-BP translation consistently outperforms PPG-to-BP translation on both Cuffless BP (Fig.~\ref{Figure3}g) and MIMIC (Supplementary Fig.~\ref{BP_Trans_Dataset} and Fig.~\ref{Figure4}k,l). Although prior studies have reported that BP estimation from ECG is generally less accurate than from PPG~\cite{finnegan2023features,landry2023current}, we hypothesize that UniCardio's multi-task, multi-modal pre-training facilitates shared representation learning across diverse task-modality combinations, enabling it to extract latent predictive features from ECG that traditional models underutilize. We defer a systematic examination of this phenomenon across broader patient cohorts and clinical contexts to future studies.
}

\begin{figure}[t]
    \centering
    \vspace{-0.1cm}
    \includegraphics[width=0.98\linewidth]{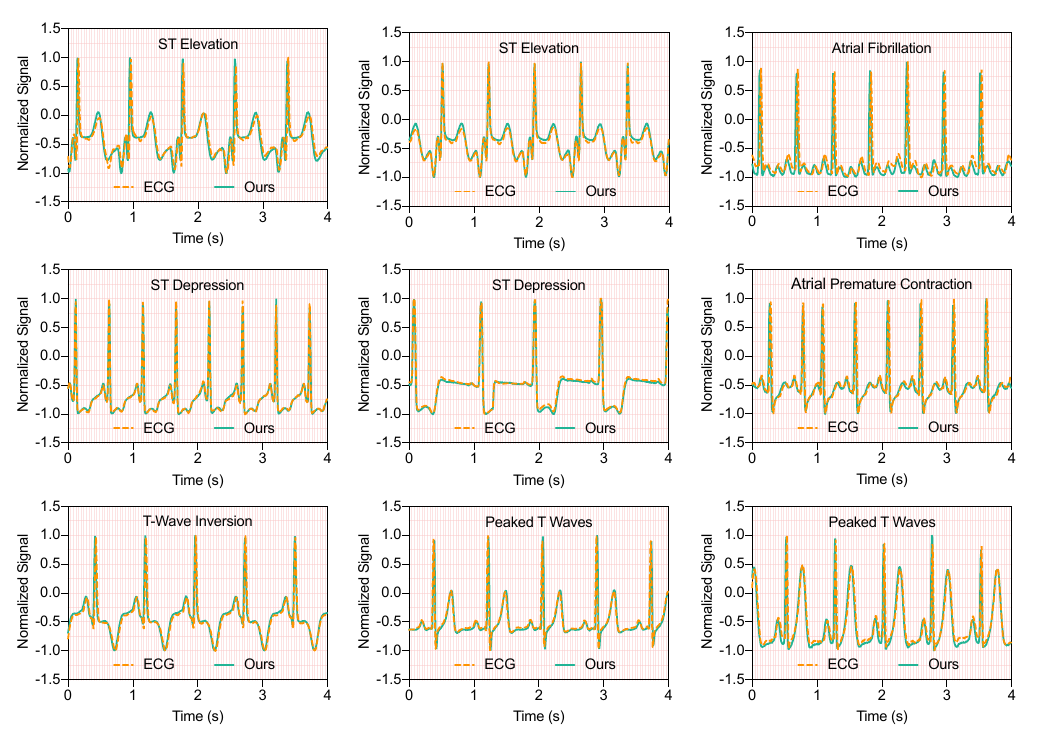}
    \caption{\textbf{Visualization of typical ECG abnormalities.}
    We visualize a range of ECG signals exhibiting typical abnormalities, alongside our generated signals from PPG-to-ECG translation. ECG grids are added to facilitate recognizing the corresponding abnormalities. The diagnostic characteristics are further validated by clinician assessments.
    } 
    \label{Figure5}
    \vspace{-0.7cm}
\end{figure}

To demonstrate UniCardio's applicability in concrete terms, we visualize a range of ECG signals exhibiting typical abnormalities, such as ST changes, biphasic P waves, T-wave inversion, peaked T waves, atrial premature contraction (APC), and AF, alongside our generated signals from PPG-to-ECG translation (Fig.~\ref{Figure5}). The generated signals align closely with the ground-truth signals, faithfully reproducing the diagnostic characteristics of corresponding abnormalities.
For example, APC is characterized by an early-occurring P wave, usually followed by a shortened PR interval and a compensatory pause before the subsequent beat, whereas AF often presents a complete absence of discrete P waves, replaced by rapid, irregular fibrillatory waves~\cite{kirchhof20162016,january20142014,zipes2017cardiac}.
Significant ST changes are critical electrocardiographic findings that may indicate acute coronary syndrome, which requires careful evaluation in conjunction with patient symptoms (e.g., chest pain), cardiac biomarkers (e.g., troponin I and tropinin T), and risk factors (e.g., smoking, obesity, and family history)~\cite{thygesen2018fourth,antman2000timi,amsterdam20142014}.
The diagnostic characteristics of these ECG signals are validated by clinician assessments (Sec.~\ref{sec44}), underscoring UniCardio's validity and interpretability. The tuning-free denoising and imputation results demonstrate similarly strong performance, with the ST change of the PTBXL dataset~\cite{wagner2020ptb} as a typical example (Supplementary Fig.~\ref{fig:denoising_st} and Fig.~\ref{fig:imputation_st}).
Besides, the diffusion process produces step-wise intermediate results that allow human experts to analyze how signals evolve throughout generation (Supplementary Fig.~\ref{fig:stepwise_diffusion} and Fig.~\ref{fig:stepwise_task}), which further enhances interpretability. 

To address the efficiency concerns commonly raised with diffusion models, we validate that UniCardio maintains comparable performance even with significantly reduced sampling steps (Supplementary Sec.~\ref{secD} and Table~\ref{tab:sampling_efficiency}). 
\textcolor{darkgreen}{
This efficient sampling results in an average inference delay of less than only 0.4 seconds per 4-second segment, well within the constraints of real-time monitoring. The accompanying pre-processing pipeline (including signal normalization, segment extraction, and entropy-based filtering) adds only 0.0034 seconds per segment on a standard CPU of the same server, rendering its overhead negligible.}
Together, these advantages establish UniCardio as a practical, robust, and interpretable tool for advancing cardiovascular healthcare.

\section{Discussion}\label{sec3}

UniCardio introduces a unified framework for multi-modal cardiovascular signal generation. Departing from traditional task-specific methods in cardiovascular signal processing, UniCardio develops a specialized diffusion transformer to model multi-modal conditional distributions of cardiovascular signals, including PPG, ECG, and BP, enabling versatile signal restoration and modality translation. By integrating modality-specific encoders and decoders, task-specific attention masks, and a continual learning paradigm, UniCardio achieves superior performance across a broad spectrum of generation tasks with diverse input-output configurations. These capabilities significantly enhance monitoring applicability and diagnostic performance for healthcare applications. The model's generative nature further allows human experts to validate the generated signals, analyze their evolution, and ensure alignment with clinical expectations. UniCardio also demonstrates remarkable efficiency in parameter costs and inference times, making it promising for deployment in wearable devices.

Cardiovascular diseases often develop silently over extended periods, with acute events posing significant health risks. Continuous monitoring in daily life is essential to detect early warning signs and facilitate timely medical intervention. UniCardio enables accurate data collection via adaptive signal restoration, especially for wearable signals that are susceptible to noise and interruptions. Additionally, certain cardiovascular signals remain inaccessible to wearable sensors, which can be synthesized by UniCardio to provide more comprehensive assessments. For patients with severe health conditions, prolonged clinical monitoring that involves non-wearable and invasive procedures can cause considerable discomfort. In such cases, modality translation offers an effective alternative for real-time alerts that prompt necessary clinical examinations. Integrating generated data with AI-driven diagnostic models further enhances automated diagnoses and offers clinicians meaningful insights, supporting proactive healthcare management.

\textcolor{darkgreen}{
Beyond healthcare, UniCardio may also benefit research in psychological and cognitive sciences, where physiological signals are widely used to assess stress~\cite{karthikeyan2011review,de2011stress}, cognitive workload~\cite{momeni2019real,ranchet2017cognitive}, and emotion recognition~\cite{shu2018review,ayata2020emotion}. In particular, pulse wave amplitude from PPG~\cite{xuan2020assessing,pavlov2023task}, HR and HR variability from ECG~\cite{al2018objective,mandrick2016neural}, as well as SBP and DBP~\cite{richter2008task,el2016rate} are common indicators of mental effort during cognitive tasks. In such non-critical care scenarios, wearable devices are typically the primary monitoring tools but often yield recordings with greater interruptions than well-controlled ICU environments. UniCardio is expected to address the recording challenges through its signal restoration capabilities. Conversely, physiological signals collected from generally healthy subjects in psychological and cognitive studies tend to exhibit higher regularity and fewer pathological features compared to ICU patient data, potentially simplifying generation tasks under certain conditions. Although empirical validation remains necessary, UniCardio's demonstrated performance holds promise for improving data quality and analytical depth in these scientific domains.
}

As a contribution to AI-generated content (AIGC)~\cite{cao2023comprehensive,mai2024brain}, UniCardio exemplifies the powerful synergy between diffusion models and transformer-based architectures in multi-modal generation, with its specialized designs effectively coordinating multi-modal relationships among versatile generation tasks. Furthermore, UniCardio highlights the importance of continual learning~\cite{wang2024comprehensive,parisi2019continual} in generative and multi-modal contexts. Traditionally, continual learning has been applied to reduce training overheads during model updates, albeit sacrificing the overall performance. Here, the proposed continual learning paradigm has proven to be a necessary approach to accommodate intricate multi-modal relationships. The proposed strategies for overcoming catastrophic forgetting allow UniCardio to continually integrate newly collected data, modalities, and tasks, ensuring its scalability and adaptability.

\textcolor{darkgreen}{
We acknowledge several limitations that warrant consideration. First, UniCardio is pre-trained on a modest amount of ICU patient data comprising three signal modalities due to the limited availability of large-scale, multi-modal public datasets. While the model achieves compelling results across a range of tasks and benchmarks, future improvements may benefit from expanded pre-training data, additional signal modalities, and annotated health conditions. Notably, UniCardio's model architecture and training paradigm inherently support continual refinement as new data and modalities become available. 
Second, while UniCardio shows promising results in handling noisy and incomplete signals, including those resembling wearable-device recordings, its current evaluations are based on benchmark datasets collected in controlled clinical settings. As such, caution is warranted when interpreting its performance for real-world wearable applications, where motion artifacts, sensor variability, and environmental interference may introduce additional challenges. These scenarios require empirical validation and potentially domain-specific adaptation.
}

\textcolor{darkgreen}{
Looking ahead, UniCardio offers a versatile and scalable foundation for multi-modal physiological signal generation, with demonstrated strengths in both signal restoration and modality translation across diverse configurations. Its model architecture and training paradigm are inherently designed to support continual integration of new datasets, signal modalities, and task demands. Future research is expected to extend UniCardio's capabilities toward broader domains of AI-assisted healthcare, including personalized monitoring, automated diagnostics, and integrated physiological data analysis.
}

\section{Methods}\label{sec4}

\subsection{Problem Formulation}\label{sec41}

Given multi-modal cardiovascular signals $\bm{s}_1, \bm{s}_2, ..., \bm{s}_k$ with a joint distribution $q(\bm{s}_1, \bm{s}_2, ..., \bm{s}_k)$, our objective is to capture all conditional distributions among these signals, which naturally covers the two main categories of cardiovascular signal processing: signal restoration and modality translation. Both categories may receive one or more cardiovascular signals as the condition modalities. 
For the sake of clarity, here we specifies $\bm{s}_i$ for $i \in [k]$ into $\bm{x},\bm{y},\bm{z}$ under $k=3$, corresponding to PPG, ECG, and BP, and their joint distribution becomes $q(\bm{x},\bm{y},\bm{z})$.

\textbf{Signal Restoration} involves two typical tasks. One is to remove the undesired information from observed recordings, such as denoising the background noise that affects the signal quality index (SQI)~\cite{elgendi2016optimal}. 
The other is synthesizing the desired information, such as the imputation of the observed intermittent signals~\cite{xu2022pulseimpute,bansal2021missing}. 
Both tasks involve transforming the low-quality data $\tilde{\bm{x}}$ into its high-quality counterpart, modeled by the conditional distribution $p(\bm{x}\vert \tilde{\bm{x}})$. When multi-modal condition information is available for signal restoration, the conditional distribution changes to $p(\bm{x}\vert \tilde{\bm{x}},\bm{y})$ or $p(\bm{x}\vert \tilde{\bm{x}},\bm{y},\bm{z})$.

\textbf{Modality Translation} refers to synthesizing signals of a target modality from off-the-shelf one(s). For example, PPG-to-ECG generation~\cite{shome2024region,sarkar2021cardiogan} and PPG-to-BP estimation~\cite{paviglianiti2020noninvasive,ibtehaz2022ppg2abp} have been separately explored to capture the conditional distribution $p(\bm{y}\vert\bm{x})$ and $p(\bm{z}\vert\bm{x})$, respectively. Similarly, when multi-modal condition information is available for modality translation, such as the BP estimation from a joint observation of PPG and ECG signals~\cite{shome2024region}, the conditional distribution changes to $p(\bm{z}\vert\bm{x},\bm{y})$. 

In summary, the restoration and translation of multi-modal cardiovascular signals can be formulated as an assembly of conditional distributions at both condition and modality levels.
Without loss of generality, we first assume the signal of target modality as $\bm{x}$ and the available cardiovascular signals are sampled from $q(\bm{x},\bm{y},\bm{z})$. 
At the condition level, we then aim to capture the modality-specific conditional distributions $p(\bm{x}\vert\bm{c_x})$, where $\bm{c_x}$ includes $2^k-1$ possibilities: (1) signal restoration $p(\bm{x}\vert\tilde{\bm{x}})$; (2) cross-modality translation $p(\bm{x}\vert\bm{y})$ and $p(\bm{x}\vert\bm{z})$; (3) multi-modal signal restoration $p(\bm{x}\vert\tilde{\bm{x}},\bm{y})$, $p(\bm{x}\vert\tilde{\bm{x}},\bm{z})$, and $p(\bm{x}\vert\tilde{\bm{x}},\bm{y},\bm{z})$; and (4) multi-modal signal translation $p(\bm{x}\vert\bm{y},\bm{z})$.
At the modality level, each signal of target modality (e.g., $\bm{x}$, $\bm{y}$, or $\bm{z}$) corresponds to the $2^k -1$ possibilities of conditional distributions, expanding the total number of tasks to $k\times(2^k-1)$ with $p(\bm{x}\vert\bm{c_x})$, $p(\bm{y}\vert\bm{c_y})$, and $p(\bm{z}\vert\bm{c_z})$.

\subsection{Generative Framework}\label{sec42}

\textbf{Diffusion Models.}
To capture the multi-modal conditional distributions inherent in cardiovascular signals, we adopt diffusion models~\cite{DDPM2020,SGMs2021} that offer distinct advantages over conventional mapping-based neural networks~\cite{zhu2019electrocardiogram,mazumder2022synthetic,ezzat2024ecg}. Diffusion models are composed of two processes.
In model training, a forward process gradually transforms the data distribution $p(\bm{x}_0)$ into a known prior distribution $p(\bm{x}_T)$, typically a standard Gaussian distribution $\mathcal{N}(\bm{0},\bm{I})$, with a sufficiently large number of forward time steps $T$. 
Among the data $\bm{x}_0$ and the noise prior $\bm{x}_T$, the time-dependent intermediate representations $\bm{x}_t$ are the noisy versions of data with increasing noise scales, which can be constructed with unconditional Gaussian transition kernel $q(\bm{x}_t\vert\bm{x}_{t-1})$:
\begin{equation}
\label{eq:forwardtext}
    q(\bm{x}_{1:T}\vert\bm{x}_0)=\prod_{t=1}^Tq(\bm{x}_t\vert\bm{x}_{t-1}),\quad p_0(\bm{x}_0)\sim p_{data}(\bm{x}).
\end{equation}
In model inference, a reverse process starts from the prior distribution $p(\bm{x}_T)$ and reconstructs the data distribution $p(\bm{x}_0)$ with iterative denoising steps. 
Each inference step $p(\bm{x}_{t-1}\vert\bm{x}_{t})$ is a mirror of the corresponding forward step $q(\bm{x}_t\vert\bm{x}_{t-1})$, and can be guided with provided condition information $\bm{c}$:
\begin{equation}
\label{eq:reversetext}
    p_{\bm{\theta}}(\bm{x}_{0:T-1}\vert\bm{x}_T,\bm{c})=p(\bm{x}_T)\prod_{t=1}^Tp_{\bm{\theta}}(\bm{x}_{t-1}\vert\bm{x}_{t},t,\bm{c}),\quad p_T(\bm{x}_T)\sim \mathcal{N}(\bm{0},\bm{I}).
\end{equation}

The generative framework of diffusion models provides two key advantages for UniCardio. 
First, because the forward process is unconditional, it enables to generate the distributions of different signal modalities $p(\bm{x})$, $p(\bm{y})$, and $p(\bm{z})$ from the same prior distribution $\mathcal{N}(\bm{0},\bm{I})$, which supports the modeling of multiple cardiovascular signals with a unified noise-to-data generation process.
Second, because the reverse process can be conditional, it allows the model to perform specific generation tasks from different conditional distributions, such as $p(\bm{x} \vert \bm{c}_x)$. 
Collectively, the unconditional forward process and the conditional reverse process can capture the distributions at both condition and modality levels, unifying multi-modal signal restoration and translation within a single framework. Further details are provided in Supplementary Sec.~\ref{secA}.

\textbf{Multi-Modal Generative Modeling.}
Unlike previous customized diffusion models designed for specific tasks in cardiovascular signal generation, such as RDDM~\cite{shome2024region} for PPG-to-ECG translation, UniCardio captures a variety of conditional distributions at both condition and modality levels. 
While recent methods~\cite{bao2023one} demonstrate success in fitting conditional distributions for text and image data, they often rely on modality-specific time steps $t^x$ for diffusion models shown in Eq.~\eqref{eq:forwardtext}.
For example, PPG-to-ECG generation would require learning the distribution $p_{\bm{\theta}}(\bm{y}_{t-1}\vert\bm{x}_{t^x},t^x, \bm{y}_{t^y},t^y,\bm{z}_{t^z},t^z)$, where $t^x=0$, $\bm{x}_{t^x}=\bm{x}_{0}$, $t^z=T$, $\bm{z}_{t^z}=\bm{z}_{T}$, in order to define the condition information as clean PPG signal $\bm{x}_{0}\sim p(\bm{x})$ without BP signal $\bm{z}_{T}\sim \mathcal{N}(\bm{0},\bm{I})$.

Moreover, UniCardio scales to more signal modalities, e.g., at least three modalities (PPG, ECG, and BP) versus two modalities (text and image). 
Providing modality-specific time steps to diffusion models combinatorially increases the complexity of generation tasks, resulting in limited training efficiency and overall performance. 
To address this, we propose a \textcolor{darkgreen}{specialized} design that simplifies the several modality-specific time steps~\cite{bao2023one} to a unified time step for all modalities, e.g., from $p_{\bm{\theta}}(\bm{y}_{t-1}\vert\bm{x}_{t^x},t^x, \bm{y}_{t^y},t^y,\bm{z}_{t^z},t^z)$ to $p_{\bm{\theta}}(\bm{y}_{t-1}\vert\bm{x}_{t}, \bm{y}_{t},\bm{z}_{t},t)$. This unified time step enables a seamless integration of multi-modal cardiovascular signal restoration and translation, as shown in Eq.~\eqref{eq:forwardtext} and Eq.~\eqref{eq:reversetext}.
Further details are provided in Supplementary Sec.~\ref{secB}.

\subsection{Model Architecture}\label{sec43}

\textbf{Modality-Specific Encoders with Multi-Scale Convolution.} 
Cardiovascular signals such as PPG, ECG, and BP are inherently complex and exhibit diverse temporal patterns that vary across different physiological states. 
When generating one of them from others, the correspondence between these modalities essentially spans multiple time scales. To handle the multi-frequency components, we design a multi-scale convolutional encoder for each signal modality, capturing features at multiple levels of temporal granularity.

Specifically, the input signals $\bm{s}_1, \bm{s}_2, ..., \bm{s}_k$ correspond to $k$ distinct modalities, where each signal $\bm{s}_i$ for $i \in [k]$ is a 1D time series of shape $(B, L, 1)$. Here, $B$ denotes the batch size and $L$ denotes the signal length. These signals are processed by modality-specific encoders $E_i(\cdot)$ for $i \in [k]$, which consist of multiple 1D CNNs with various kernel sizes $\{1, 3, 5, 7, 9, 11\}$ to extract features at different temporal scales. The outputs from the kernels are concatenated along the feature dimension, producing a feature vector of shape $(B, L, C)$, where $C$ represents the number of channels. 
The feature vectors from all modalities are concatenated along the temporal dimension into a joint feature vector of shape $(B, k L, C)$:
\begin{equation}
    \bm{h}_s = [\bm{h}_{s,1}; \bm{h}_{s,2}; ... ; \bm{h}_{s,k}],
\end{equation}
where $\bm{h}_{s,i} = E_i(\bm{s}_i)$ for $i \in [k]$.
We further introduce an auxiliary modality (AM) as a placeholder for signal restoration tasks where the target modality is occupied as a condition modality.
The input signals for AM are non-informative, serving as the start of diffusion process to facilitate the generative modeling, as will be detailed later. 
For notation clarity, we update $k \leftarrow k+1$ in subsequent descriptions, with $k=4$ corresponding to PPG, ECG, BP, and AM.

\textbf{Customized Transformer Modules.}
The joint feature vector $\bm{h}_s$ is processed through multiple customized transformer modules (Supplementary Fig.~\ref{fig:model_architecture}), which facilitate intra- and inter-modal interactions among signal points at different timestamps. For clarity, we denote the input to a module as $\bm{h}$ ($\bm{h} = \bm{h}_s$ for the first module) and its output as $\bm{h}^{'}$, omitting the module identity where unnecessary.
In each module, the diffusion embedding $\bm{h}_d$ is first added to the input feature vector, updating $\bm{h} \leftarrow \bm{h} + \bm{h}_d$. The updated vector $\bm{h}$ is then passed to a multi-head self-attention (MSA) layer. In order to control the involved modalities in different generation tasks, each MSA layer employs a task-specific attention mask $M$ of shape $(kL, kL)$, which constrains the information flow between condition and target modalities. Specifically, the key $\bm{h}_K$, query $\bm{h}_Q$ and value $\bm{h}_V$ are computed as linear transformation of $\bm{h}$:
\begin{equation}
\bm{h}_K = \bm{h} W_K, \, \bm{h}_Q = \bm{h} W_Q, \, \bm{h}_V = \bm{h} W_V,
\end{equation}
where $W_K \in \mathbb{R}^{C \times d_K}$, $W_Q \in \mathbb{R}^{C \times d_Q}$, and $W_V \in \mathbb{R}^{C \times d_V}$ are learnable matrices. Here, $d_K$, $d_Q$, and $d_V$ denote the dimensions of the key, query, and value, respectively, with $d_K = d_Q$. Then, we implement $M$ into the self-attention operation as follows:
\begin{equation}
\bm{h}^{'} = \text{MSA}(\bm{h}, M) = \text{Softmax}\left( \frac{\bm{h}_Q \bm{h}_K^\top}{\sqrt{d_K}} + M \right) \bm{h}_V,
\end{equation}
where $\bm{h}^{'}$ and $\bm{h}$ keep the same shape $(B, kL, C)$.

The task-specific attention mask $M$ controls the information flow with predefined values. Each modality $i \in [k]$ maps to the token range $[(i-1)L : iL]$ of $\bm{h}$ (Supplementary Table~\ref{tab:attention}).
For translation tasks from modality $i$ to modality $j$, $M$ permits intra-modal interactions within $i$, intra-modal interactions within $j$, and inter-modal interactions from $i$ to $j$. The elements of $M$ corresponding to these interactions are set to zero, while other elements are assigned large negative values to block irrelevant interactions during the softmax operation.
For denoising or imputation tasks, where AM acts as the target modality ($j = k$), $M$ similarly ensures information flow of intra-modal interactions within $i$, intra-modal interactions within $k$, and inter-modal interactions from $i$ to $k$, while blocking other irrelevant interactions. 
This unified masking mechanism enables consistent handling of generative tasks involving different modalities and input-output configurations.

The output $\bm{h}^{'}$ from the MSA layer is refined by a fully connected (FC) layer $F_{1}(\cdot)$ and combined with the time embedding $\bm{h}_t$, updating $\bm{h}^{'} \leftarrow F_1(\bm{h}^{'}) + \bm{h}_t$. This operation expands the feature dimension, resulting in a feature vector of shape $(B, kL, 2C)$. We then implement a gated activation unit~\cite{van2016conditional,van2016wavenet} to capture the complex conditional distributions of signal modalities given the updated feature vector $\bm{h}^{'}$. Specifically, we split $\bm{h}^{'}$ along the feature dimension into two parts, i.e., $\bm{h}^{'} = [\bm{h}^{'}_1; \bm{h}^{'}_2]$, each of shape $(B, kL, C)$. These two parts are processed separately through $\tanh$ and $\sigma$ activations, respectively, and combined element-wise as $\bm{h}' \leftarrow \tanh(\bm{h}'_1) \odot \sigma(\bm{h}'_2)$. This operation produces a transformed feature vector $\bm{h}^{'}$ with a reduced dimension $(B, kL, C)$.

We further implement residual and skip connections~\cite{kongdiffwave,tashiro2021csdi} across customized transformer modules to facilitate convergence of the entire model. Specifically, the transformed feature vector $\bm{h}^{'}$ is projected back to an expanded dimension of $(B, kL, 2C)$ though another FC layer $F_{2}(\cdot)$ and split again into two parts, $\bm{h}^{'} = [\bm{h}^{'}_1; \bm{h}^{'}_2]$, each of shape $(B, kL, C)$. The first part $\bm{h}^{'}_1$ is added to the input of the current transformer module, serving as the input to the next transformer module. The second part $\bm{h}^{'}_2$ is accumulated across all transformer modules and passed through an additional FC layer $F_3(\cdot)$. This produces the final feature vector $\bm{h}^{'}_{s} = F_3(\sum \bm{h}^{'}_{2})$, retaining the shape $(B, kL, C)$.

\textbf{Modality-Specific Decoders.} 
The final feature vector $\bm{h}^{'}_{s}$ is split into modality-specific components $\bm{h}^{'}_{s,1}, \bm{h}^{'}_{s,2}, ..., \bm{h}^{'}_{s,k}$, each of shape $(B, L, C)$. These components are processed through modality-specific decoders $D_i(\cdot), i \in [k-1]$, where each decoder is implemented as a two-layer MLP with ReLU activation. The decoders project the split feature vectors into their respective generated signals of shape $(B, L, 1)$:
\begin{equation}
    \hat{\bm{s}}_i = D_i(\bm{h}^{'}_{s,i}), i \in [k].
\end{equation}
In particular, the first $k-1$ decoders $D_i, i \in [k-1]$ are optimized from network training, whereas the last decoder $D_k$ of AM inherits weights from one of the $k-1$ decoders depending on the generation tasks. For example, when performing ECG imputation with $k=4$ representing PPG, ECG, BP and AM, the decoder $D_{k}$ reuses the weights of $D_{2}$ to generate ECG signals. This strategy ensures consistency across tasks and reduces the need for additional training of the final decoder.

\subsection{Experimental Setup}\label{sec44}

\textbf{Dataset.} UniCardio is pre-trained with the Cuffless BP dataset~\cite{kachuee2016cuffless}, which contains 339 hours of trimodal recordings (PPG, ECG, and BP) collected from ICU patients. 
PPG and ECG signals are processed using a 0.5 Hz high-pass Butterworth filter (order = 5) and a 50 Hz pulse filter, followed by z-score normalization \cite{makowski2021neurokit2}. BP signals are left unfiltered and unnormalized to preserve their absolute magnitude. 
\textcolor{darkgreen}{
To ensure data quality, we address recording errors such as sensor displacements and device disconnections by filtering out such signals based on sample entropy~\cite{richman2000physiological}, which quantifies the regularity and complexity of time series by estimating the likelihood that similar patterns remain consistent over time. We empirically determined entropy thresholds by analyzing the distribution of values across the training set and qualitatively inspecting corresponding signal segments. Specifically, we found that segments with entropy values above 0.2 for PPG and BP, and above 0.3 for ECG, typically exhibit excessive irregularity or noise and lack the expected periodic or quasi-periodic structure. These thresholds effectively remove physiologically implausible segments while preserving sufficient high-quality data for training and evaluation.}
ECG signals are more prone to corruption due to their high-frequency components and therefore require additional pre-processing. We first detect and rectify inverted ECG signals caused by sensor displacements, and then apply SQI-based selection~\cite{makowski2021neurokit2} for further refinements. Finally, PPG and ECG signals are min-max normalized, while BP signals are z-scored using the mean and standard deviation calculated from the training set. The processed signals are segmented into 4-second pairs and randomly split into training, validation, and test sets in an 80\%-10\%-10\% ratio.

\textbf{Training Regime.} 
For continual learning, the entire training process (a total of $e$ epochs) is divided into $k$ phases ($e/k$ epochs per phase), incorporating generation tasks conditioned on an increasing number of modalities. Corresponding to the three cardiovascular signals, i.e., PPG, ECG, and BP, we have a total of four phases ($k=4$). The first three phases are used to learn the generation tasks conditioned on one, two, and three modalities, respectively, while the last one is used to balance the task distribution. 
In the first phase, the model is trained exclusively on one-condition tasks, with training batches equally allocated between translation and imputation for a single condition modality.
In the second phase, two-condition tasks are introduced. The training batches for one- and two-condition tasks are 50\% each, equally allocated between translation and imputation.
In the third phase, three-condition tasks are incorporated. The training batches allocate 25\% to one- and two-condition tasks of translation and imputation, and 50\% to three-condition tasks of imputation only.
In the final phase, the model is fine-tuned to balance the one-, two-, and three-condition tasks, allocated equal proportions of training batches.

The task-specific attention masks control the condition modalities for each generation task throughout the training process. The model is optimized using an SGD optimizer over the total $e$ epochs. The learning rate starts as $1\times10^{-3}$ at the beginning of the first $e/k$ epochs to provide a strong initialization for one-condition tasks. It is then reduced to $1\times10^{-4}$ at epoch $0.7\times e/k$ as the condition modalities increases. In the final $e/k$ epochs, the learning rate is further decreased to $1\times10^{-5}$ to enable precise fine-tuning and ensure balanced performance across all tasks. We empirically identify that $e=800$ already achieves superior performance. Further details are provided in Supplementary Sec.~\ref{secC}, and the pseudo codes are provided in Supplementary Sec.~\ref{secE}. The entire model is pre-trained with 8-card RTX 4090 GPUs for around 7 days. The carbon footprint is around 225kg CO$_2$.

\textbf{Downstream Application.}
After the pre-training stage, we consider a variety of downstream applications on different datasets, including the PTBXL dataset~\cite{wagner2020ptb} for detection of abnormal health conditions, the MIMIC PERform AF dataset~\cite{charlton2022detecting} for detection of AF, the WESAD dataset~\cite{schmidt2018introducing} for HR estimation, and the MIMIC dataset~\cite{moody1996database} for SBP/DBP estimation. These datasets are processed with a similar pipeline as the pre-training dataset. 
The detection of AF, ST change, and hypertrophy is performed by training respective classification models based on 1D VGG-16 architectures~\cite{liu2022multiclass}. HR estimation is conducted by detecting heartbeat peaks with common algorithms~\cite{nabian2018open}. 
SBP/DBP estimation is conducted by training a regression model based on a CNN-LSTM architecture~\cite{jeong2021combined}.

\textbf{Evaluation Metric.} 
We evaluate the quality of generated signals with three common metrics, including the root mean squared error (RMSE), mean absolute error (MAE), and Kolmogorov-Smirnov test (KS-Test). 
\textcolor{darkgreen}{To account for small temporal misalignments that may disproportionately affect pointwise metrics, we further report the minimum RMSE (Min-RMSE) and minimum MAE (Min-MAE) based on bounded temporal shifts. These metrics quantify the best achievable reconstruction error after applying small temporal shifts to generated signals, providing a more robust assessment of morphological similarity.}
Moreover, we assess the ability to detect cardiovascular abnormalities with accuracy, specificity, and sensitivity, which reflect different proportions of true positives (TP), true negatives (TN), false positives (FP), and false negatives (FN).

\begin{itemize}

\item \textbf{RMSE} quantifies the square root of the mean of the squared differences between the generated signal $\hat{\bm{x}}$ and the ground-truth signal $\bm{x}$:  
\begin{equation}  
    \mathrm{RMSE} = \sqrt{\frac{1}{N} \sum_{i=1}^{N} (\hat{\bm{x}}_i - \bm{x}_i)^2},  
\end{equation}  
where $N$ is the total number of samples.  

\item \textbf{MAE} evaluates the average absolute differences between the generated signal $\hat{\bm{x}}$ and the ground-truth signal $\bm{x}$:  
\begin{equation}  
\mathrm{MAE} = \frac{1}{N} \sum_{i=1}^{N} \vert\hat{\bm{x}}_i - \bm{x}_i\vert.  
\end{equation}  

\textcolor{darkgreen}{
\item \textbf{Min-RMSE} quantifies the lowest achievable RMSE between generated and ground-truth signals after applying small temporal shifts to account for misalignments:
\begin{equation}
\mathrm{Min\text{-}RMSE} = \min_{\tau \in \mathcal{T}} \sqrt{\frac{1}{N} \sum_{i=1}^{N} \left(\hat{\bm{x}}^{(\tau)}_i - \bm{x}_i\right)^2},
\end{equation}
where $\mathcal{T}$ denotes a finite set of allowable temporal shifts (within one second in our case), and $\hat{\bm{x}}^{(\tau)}_i$ correspond to $\hat{\bm{x}}_i$ shifted by $\tau$ signal points. 
}

\textcolor{darkgreen}{
\item \textbf{Min-MAE} similarly reports the minimum achievable MAE after applying small temporal shifts:
\begin{equation}
\mathrm{Min\text{-}MAE} = \min_{\tau \in \mathcal{T}} \frac{1}{N} \sum_{i=1}^{N} \vert \hat{\bm{x}}^{(\tau)}_i - \bm{x}_i \vert.
\end{equation}
}

\item \textbf{KS-Test} evaluates the maximum distance between the cumulative distribution functions of the generated signal $F_{\hat{\bm{x}}}(a)$ and the ground-truth signal $F_{\bm{x}}(a)$, with the supremum over all possible values of $a$:  
\begin{equation}  
\mathrm{KS\text{-}Test} = \sup_{a} \vert F_{\hat{\bm{x}}}(a) - F_{\bm{x}}(a) \vert. 
\end{equation}  

\item \textbf{Accuracy} measures the rate of correctly classified samples, defined as:  
\begin{equation}  
\mathrm{Accuracy} = \frac{\mathrm{TP} + \mathrm{TN}}{\mathrm{TP} + \mathrm{TN} + \mathrm{FP} + \mathrm{FN}}.
\end{equation}  

\item \textbf{Specificity} quantifies the true negative rate, defined as:  
\begin{equation}  
\mathrm{Specificity} = \frac{\mathrm{TN}}{\mathrm{TN} + \mathrm{FP}}.  
\end{equation}  

\item \textbf{Sensitivity} quantifies the true positive rate, defined as:  
\begin{equation}  
\mathrm{Sensitivity} = \frac{\mathrm{TP}}{\mathrm{TP} + \mathrm{FN}}.  
\end{equation}  

\end{itemize}

\textbf{Clinician Assessment.} The diagnostic characteristics of ground-truth and generated ECG signals are validated by two cardiology clinicians in a back-to-back manner (a third clinician is asked to verify in case of inconsistency) for routine quality assurance. The employed datasets~\cite{kachuee2016cuffless} are fully anonymized, de-identified and have been previously released for research purposes. We confirmed with the Tsinghua University Ethics Committee that such routine analysis and processing of publicly available data does not require special approval.

\backmatter

\section*{Data Availability}
All benchmark datasets used in this paper are publicly available, including the Cuffless BP dataset~\cite{kachuee2016cuffless}, the PTBXL dataset~\cite{wagner2020ptb}, the MIMIC PERform AF dataset~\cite{charlton2022detecting}, and the WESAD dataset~\cite{schmidt2018introducing}.

\section*{Code Availability}
The implementation code is included in Supplementary Materials for examination, and will be released at \url{https://github.com/thu-ml/UniCardio} upon acceptance.

\section*{Acknowledgments}
This work was supported by the NSFC Projects (62406160, 62350080, 92270001, U24A20342), Tsinghua Institute for Guo Qiang, and the High Performance Computing Center, Tsinghua University. J.Z. is also supported by the XPlorer Prize.

\section*{Author Contributions Statement}
Z.C., Y.M., L.W. and J.Z. conceived the project. 
Z.C., Y.M. and L.W. designed the computational model. 
Y.M. performed main experiments, assisted by Z.C. and L.W..
Z.C., Y.M. and L.W. analyzed the data.
L.W. wrote the paper, assisted by Z.C., Y.M. and L.F..
All authors revised the paper.
L.W. and J.Z. supervised the project. 

\section*{Competing Interests Statement}
The authors declare no competing interests.


\bibliography{sn-bibliography}


\clearpage
\appendix


\setcounter{table}{0}
\setcounter{figure}{0}

\clearpage
\section{Diffusion Models}\label{secA}

\subsection{Unconditional Diffusion Models}
\label{secA1}

Diffusion models, known as denoising diffusion probabilistic models~\cite{DDPM2020} or score-based generative models~\cite{SGMs2021}, approximate data distributions by learning time-dependent score functions that reverse a predefined data-to-noise forward process. During sampling, they iteratively remove Gaussian noise and refine the generation results into structured data through a coarse-to-fine trajectory, enabling high-fidelity generation across diverse modalities such as image~\cite{bao2023one,SD1CVPR2022,SD3ICML2024,UViT2023}, audio~\cite{InferGrad2022,AudioLDM2023,DiffGAP2025,TiVA2024}, 3D shape~\cite{ProlificDreamer}, and video~\cite{Vidu,FrameBridge}. Owing to their strong generative performance, diffusion models have become foundational components in modern data generation systems.

Recent studies have extended diffusion models to bio-electrical signal processing~\cite{RespDiff2025}, including applications in pulsative signals imputation and forecasting~\cite{PulseDiff,xu2022pulseimpute,BPForecasting} as well as modality translation such as PPG-to-ECG synthesis~\cite{shome2024region}. 
However, most previous studies remain task-specific, lacking a unified framework for multi-modal signal generation. Despite the growing importance of cardiovascular monitoring, a diffusion model capable of handling multiple signal modalities within a unified system remains underexplored.

Diffusion models comprise two coupled Markov chains: a \textit{forward process} that progressively transforms data distribution into a Gaussian noise distribution, and a \textit{reverse process} that iteratively reconstructs data from the noise. 
The forward process defines a transformation from a data sample $\bm{x}_0 \sim p_{\text{data}}(\bm{x})$ to a noisy latent $\bm{x}_T$, governed by a predefined noise schedule over $T$ time steps:
\begin{equation}
\label{eq:appforward}
    q(\bm{x}_{1:T} \vert \bm{x}_0) = \prod_{t=1}^{T} q(\bm{x}_t \vert \bm{x}_{t-1}), \quad q_0(\bm{x}_0) \sim p_{\text{data}}(\bm{x}).
\end{equation}
The transition kernel $q(\bm{x}_t \vert \bm{x}_{t-1})$ is typically defined as:
\[
q(\bm{x}_t \vert \bm{x}_{t-1}) = \mathcal{N}(\bm{x}_t; \sqrt{1 - \beta_t} \bm{x}_{t-1}, \beta_t \bm{I}),
\]
where $\beta_t$ is a small positive constant. For efficiency, the marginal distribution at time step $t$ can be directly computed as:
\begin{equation}
\label{eq:forward}
    q(\bm{x}_t \vert \bm{x}_0) = \mathcal{N}(\bm{x}_t; \sqrt{\bar{\alpha}_t} \bm{x}_0, (1 - \bar{\alpha}_t)\bm{I}),
\end{equation}
where $\alpha_t = 1 - \beta_t$ and $\bar{\alpha}_t = \prod_{s=1}^{t} \alpha_s$.

The reverse process starts from a standard Gaussian prior $p_T(\bm{x}_T)$ and aims to reconstruct the data distribution with iterative sampling steps:
\begin{equation}
\label{eq:appreverse}
    p_{\bm{\theta}}(\bm{x}_{0:T-1} \vert \bm{x}_T) = p(\bm{x}_T)\prod_{t=1}^{T} p_{\bm{\theta}}(\bm{x}_{t-1} \vert \bm{x}_t), \quad p(\bm{x}_T) \sim \mathcal{N}(\bm{0}, \bm{I}).
\end{equation}
Each reverse transition is parameterized as:
\begin{equation}
\label{eq:appsinglereverse}
    p_{\bm{\theta}}(\bm{x}_{t-1} \vert \bm{x}_t) = \mathcal{N}(\bm{x}_{t-1}; \bm{\mu}_{\bm{\theta}}(\bm{x}_t, t), \sigma^2_{\bm{\theta}}(\bm{x}_t, t)\bm{I}),
\end{equation}
where $\bm{\mu}_{\bm{\theta}}$ and $\sigma^2_{\bm{\theta}}$ are the learned mean and variance functions. In practice, the variance $\sigma_t^2$ is often fixed as $\sigma_t^2 = \frac{1 - \bar{\alpha}_{t-1}}{1 - \bar{\alpha}_t}$ or simply $\sigma_t^2 = \beta_t$, which yields similar performance while simplifying training~\cite{DDPM2020,DDIM,kongdiffwave}.

The mean function $\bm{\mu}_{\bm{\theta}}$ can be derived from various parameterizations such as score, noise, or data prediction. Without loss of generality, we adopt noise prediction and therefore have:
\begin{equation}
    \bm{\mu}_{\bm{\theta}}(\bm{x}_t, t) = \frac{1}{\sqrt{\alpha_t}} \left( \bm{x}_t - \frac{\beta_t}{\sqrt{1 - \bar{\alpha}_t}} \bm{\epsilon}_{\bm{\theta}}(\bm{x}_t, t) \right).
\end{equation}
Then, the training objective of diffusion models minimizes the prediction error of the noise added in the forward process:
\begin{equation}
\label{eq:TrainingObjective}
    \mathcal{L}_{\text{udm}}(\bm{\theta}) = \mathbb{E}_{\bm{x}_0, \bm{\epsilon}, t} \left[ \left\| \bm{\epsilon} - \bm{\epsilon}_{\bm{\theta}}\left( \sqrt{\bar{\alpha}_t} \bm{x}_0 + \sqrt{1 - \bar{\alpha}_t} \bm{\epsilon}, t \right) \right\|_2^2 \right].
\end{equation}

\subsection{Conditional Diffusion Models}
\label{secA2}

In many tasks such as signal restoration or modality translation tasks, condition information $\bm{c}$ is available and can be leveraged to guide the generative process. For instance, PPG-to-BP estimation uses observed PPG signals as the conditional input to indicate the generation of continuous BP waveforms~\cite{DiffCNBP}. 

Given conditioning inputs $\bm{c}$, the training objective of conditional diffusion models becomes:
\begin{equation}
\label{eq:appcdmTrainingObjective}
    \mathcal{L}_{\text{cdm}}(\bm{\theta}) = \mathbb{E}_{\bm{x}_0, \bm{\epsilon}, \bm{c}, t} \left[ \left\| \bm{\epsilon} - \bm{\epsilon}_{\bm{\theta}}(\bm{x}_t, t, \bm{c}) \right\|_2^2 \right],
\end{equation}
where $\bm{c}$ is usually taken as the model input. 
The corresponding sampling process becomes:
\begin{equation}
\label{eq:appcdmreverse}
    p_{\bm{\theta}}(\bm{x}_{0:T-1} \vert \bm{x}_T, \bm{c}) = p(\bm{x}_T)\prod_{t=1}^{T} p_{\bm{\theta}}(\bm{x}_{t-1} \vert \bm{x}_t, \bm{c}), \quad p(\bm{x}_T) \sim \mathcal{N}(\bm{0}, \bm{I}),
\end{equation}
\begin{equation}
     p_{\bm{\theta}}(\bm{x}_{t-1} \vert \bm{x}_t, \bm{c}) = \mathcal{N}(\bm{x}_{t-1}; \bm{\mu}_{\bm{\theta}}(\bm{x}_t, t, \bm{c}), \sigma_t^2 \bm{I}),
\end{equation}
where the mean function is guided by the learned conditional noise estimation:
\begin{equation}
    \bm{\mu}_{\bm{\theta}}(\bm{x}_t, t, \bm{c}) = \frac{1}{\sqrt{\alpha_t}} \left( \bm{x}_t - \frac{\beta_t}{\sqrt{1 - \bar{\alpha}_t}} \bm{\epsilon}_{\bm{\theta}}(\bm{x}_t, t, \bm{c}) \right).
\end{equation}

When multiple condition modalities $\bm{c}_1, \dots, \bm{c}_k$ are available (e.g., ECG and PPG used jointly for BP estimation), they can be incorporated into a multi-conditional generative process. The training objectives and conditional sampling process become:
\begin{equation}
    \mathcal{L}_{\text{mcdm}}(\bm{\theta}) = \mathbb{E}_{\bm{x}_0, \bm{\epsilon}, \bm{c}_1, \dots, \bm{c}_k, t} \left[ \left\| \bm{\epsilon} - \bm{\epsilon}_{\bm{\theta}}(\bm{x}_t, t, \bm{c}_1, \dots, \bm{c}_k) \right\|_2^2 \right],
\end{equation}
\begin{equation}
    p_{\bm{\theta}}(\bm{x}_{0:T-1} \vert \bm{x}_T, \bm{c}_1, \dots, \bm{c}_k) = p(\bm{x}_T)\prod_{t=1}^{T} p_{\bm{\theta}}(\bm{x}_{t-1} \vert \bm{x}_t, \bm{c}_1, \dots, \bm{c}_k), p(\bm{x}_T) \sim \mathcal{N}(\bm{0}, \bm{I}).
\end{equation}
Here, $k$ denotes the number of condition signals, and $p_T(\bm{x}_T)$ remains the unconditional Gaussian prior.

\section{Unified Generation Prior} 
\label{secB}

UniCardio is designed to accommodate generation tasks across multiple cardiovascular signals within a single conditional diffusion framework. These modalities exhibit substantial differences in temporal dynamics, amplitude scaling, and physiological semantics. To enable unified modeling under such heterogeneity, UniCardio employs a shared, uninformative Gaussian prior derived from an unconditional forward process. This prior remains agnostic to both the generation target and the conditioning configuration. As a result, UniCardio supports versatile generation of each target modality with various conditioning settings.
All tasks are performed with a unified model architecture, with task-specific sampling trajectories guided by conditional inputs and initialized from the shared latent distribution.

\paragraph{Unconditional Forward Process.}
In UniCardio, the forward process is formulated to be unconditional across all tasks. Given a clean signal $\bm{x}_0$, the process perturbs it into a noisy latent variable $\bm{x}_t$ as follows:
\[
\bm{x}_t = \sqrt{\bar{\alpha}_t} \bm{x}_0 + \sqrt{1 - \bar{\alpha}_t} \bm{\epsilon}, \quad \bm{\epsilon} \sim \mathcal{N}(\bm{0}, \bm{I}),
\]
where $\bar{\alpha}_t$ is a monotonically decreasing function of the diffusion time $t$. As $t \to 1$, the contribution from the original data diminishes, and $\bm{x}_t$ converges toward Gaussian noise. The resulting terminal distribution $p_T(\bm{x}) \approx \mathcal{N}(\bm{0}, \bm{I})$ serves as a shared latent prior across all generation tasks.
All generative trajectories, regardless of the target modality or conditioning settings, originate from this common latent distribution defined by the unconditional forward process.

\paragraph{Conditional Sampling Trajectories.}
UniCardio models different signal generation tasks through specific conditional reverse process. Given the same unconditional latent prior, each restoration or translation task is realized as a distinct conditional sampling trajectory defined by the observed inputs. The conditioning signals $\bm{c}$ guide the denoising trajectory from $\bm{x}_T$ to $\bm{x}_0$, directing generation toward the target modality while the noise-to-data generative framework remains invariant across tasks.

This separation between a task-invariant forward process and a task-specific reverse process is central to the flexibility of UniCardio. It enables the model to accommodate diverse input combinations without retraining or modifying the architecture. More importantly, the use of a shared, unconditional prior allows UniCardio to model all target modalities (e.g., PPG, ECG, and BP) within a single diffusion model, despite their distinct signal structures and semantic differences. During inference, various conditioning settings, including partially observed or missing signal modalities, can be seamlessly supported by adjusting the conditioning function alone, without altering the generative mechanism.

\paragraph{Comparison with Alternative Approaches.}
The aforementioned formulation contrasts with encoder-decoder and mapping-based architectures, where latent representations are typically entangled with the structure of the condition or target modality. Such designs often require architectural adaptation or retraining to accommodate new generation tasks or modalities.
Recent bridge models~\citep{FrameBridge,DDBM} construct task-specific priors from degraded observations to improve performance in structured domains. While effective in settings with known alignment between source and target signals, these methods rely on assumptions of structural similarity that do not hold for cardiovascular signals: PPG, ECG, and BP vary significantly in waveform morphology, sampling scale, and physiological semantics.

By maintaining an unconditional forward process and defining task-specific behavior through conditional sampling, UniCardio preserves a coherent latent space while enabling highly flexible multi-task modeling. In total, UniCardio supports 33 distinct generation tasks over 3 signal modalities, encompassing both signal restoration and modality translation. For restoration, two forms of degraded input are considered: additive noise for denoising and randomly masked observations for imputation. All tasks across PPG, ECG, and BP, and across varying conditioning configurations are unified within a \textit{single generative framework} without requiring task-specific retraining or model adaptation.

\section{Unified Training Process}
\label{secC}

UniCardio aims to model conditional distributions among PPG, ECG, and BP signals using a \textit{single model architecture} within a \textit{unified generative framework}. In diffusion models, training is inherently a form of multi-task learning~\citep{Min-SNR}, as the model must approximate data distributions conditioned on the diffusion time step $t\sim (0,T]$. Consequently, diffusion model training typically demands millions of iterations~\citep{DDPM2020,SGMs2021}, extensive computational resources~\citep{SD1CVPR2022,SD3ICML2024}, and careful optimization across varying noise scales\citep{Grad-TTS}.

Extending this complexity, UniCardio is designed to simultaneously learn across multiple target modalities and conditioning configurations. Specifically, it learns a family of \textit{multi-modal, multi-condition, and time-dependent} score functions $\bm{\epsilon}_{\theta}(\bm{x}_t,\bm{c},t)$, where the target data $\bm{x}_0$ denotes the target signal, $\bm{c}$ represents the observed signal(s) as the condition(s), and $t$ is the diffusion time step.
For efficient and high-fidelity multi-modal generation without relying on task-specific specialization, it necessitates a careful integration of conditional learning principles and a systematically structured training strategy.

To this end, we first revisit the foundational concepts of conditional learning in diffusion models, and subsequently describe the progressive training methodology adopted by UniCardio.

\subsection{Conditional Learning}
The training of diffusion models can be categorized into different modes depending on the availability of auxiliary information. Following the framework outlined in SCDM~\citep{SCDM}, we distinguish between unconditional, conditional, and multi-conditional learning settings.

\paragraph{Unconditional Learning.} In the unconditional setting, the model is trained to fit the full data distribution without access to conditioning information. 
The training objective is given by: \begin{equation} \label{appudmlearning} \min_{\bm{\theta} \in \Theta} \mathcal{D}(q(\bm{x}) \Vert p_{\bm{\theta}}(\bm{x})), \end{equation} 
where $\mathcal{D}$ denotes a divergence metric (e.g., KL divergence), $q(\bm{x})$ is the empirical data distribution, and $\bm{\theta}$ denote the model parameters.
Under this setting, the model is required to capture adequate variability inherent in the data, such as modality-specific structures and inter-subject differences. For cardiovascular signals, this involves representing critical characteristics, such as the dicrotic notch in PPG waveforms, the QRS complex in ECG signals, and systolic and diastolic peaks in BP waveforms, while accommodating typical temporal dynamics and variability across individuals.

\paragraph{Conditional Learning.} Conditional learning introduces auxiliary information $\bm{c}$ into the modeling process. An embedding function $\phi \in \Phi$ transforms the condition into a feature representation, leading to the following training objective: \begin{equation} \label{appcdmlearning} \min_{\bm{\theta} \in \Theta, \phi \in \Phi} \mathbb{E}{q(\bm{c})} \left[ \mathcal{D}(q(\bm{x} \vert \bm{c}) \Vert p_{\bm{\theta,\phi}}(\bm{x} \vert \bm{c})) \right]. \end{equation}

By conditioning on $\bm{c}$, the generative task is restricted to a subset of the data distribution, typically resulting in a narrower and less complex target.
For instance, generating BP signals conditioned on observed PPG signals focuses the model's capacity on more homogeneous temporal and morphological patterns, enabling more efficient learning and higher-quality synthesis relative to the unconditional case. 

\paragraph{Multi-Conditional Learning.} In clinical applications, multiple sources of conditional information may be available simultaneously. The training objective generalizes accordingly: \begin{equation} \label{appcdmTrainingObjective} \min_{\bm{\theta} \in \Theta, \phi \in \Phi} \mathbb{E}_{q(\bm{c}_1, \dots, \bm{c}_k)} \left[ \mathcal{D}(q(\bm{x} \vert \bm{c}_1, \dots, \bm{c}_k) \Vert p_{\bm{\theta,\phi}}(\bm{x} \vert \bm{c}_1, \dots, \bm{c}_k)) \right]. \end{equation}

Leveraging multiple conditions further constrains the generative space, thereby simplifying the training process and improving generation fidelity.
For example, BP estimation conditioned jointly on PPG and ECG signals yields superior performance compared to using a single modality alone~\citep{kachuee2016cuffless,BeatFiltering}.

\subsection{Progressive Training of UniCardio}

Building on the principles of conditional learning, UniCardio adopts a structured training strategy to facilitate efficient multi-modal, multi-condition generation of cardiovascular signals. Rather than training task-specific models, UniCardio progressively handles increasingly constrained conditional distributions within a unified framework.
Motivated by the observation that additional conditions simplify the generative target, we design a continual learning paradigm with adaptive training batch composition and learning rate scheduling (a detailed pseudo-code is provided in Algorithm~\ref{alg:unified-training}). The entire training process is divided into $k=4$ sequential phases over a total of $e=800$ epochs, with each phase lasting $e/k$ epochs:

Phase 1: The model is trained exclusively on one-condition tasks, with training batches equally split between translation and imputation.

Phase 2: Two-condition tasks are introduced. The training batches allocate 50\% to one- and two-condition tasks, equally distributed between translation and imputation.

Phase 3: Three-condition tasks are introduced. The training batches allocate 25\% to one- and two-condition tasks of both translation and imputation, and 50\% to three-condition tasks of imputation only.

Phase 4: A balanced fine-tuning stage. The training batches allocate equal proportions to one-, two-, and three-condition tasks.

Throughout training, task-specific attention masks control the visibility of condition and target modalities, enabling dynamic cross-modal learning. 
The learning rate starts at $1\times 10^{-3}$ during the first $e/k$ epochs to provide a strong initialization for one-condition tasks. It is then reduced to $1\times 10^{-4}$ at epoch $0.7\times e/k$ as the condition modalities increase. In the final $e/k$ epochs, the learning rate is further decreased to $1\times 10^{-5}$ to enable precise fine-tuning and ensure balanced performance across all tasks.



\section{Efficient Sampling Process}
\label{secD}

\subsection{First-Order ODE Sampler}

Despite the strong generative capabilities, diffusion models suffer from inherently slow inference speed. Traditional samplers based on Langevin dynamics require a large number of denoising steps to gradually transform noise into a high-quality sample. While reducing the number of steps can accelerate sampling, it often leads to severe degradation in synthesis quality. Therefore, designing an efficient sampling algorithm that preserves generation fidelity while reducing inference cost is critical for the practical deployment of diffusion models, particularly in modeling cardiovascular signals where long-duration waveforms such as PPG, ECG, and BP traces demand both high temporal resolution and morphological accuracy.

To address this limitation, UniCardio employs Denoising Diffusion Implicit Models (DDIM)~\citep{DDIM}, a training-free first-order ordinary differential equation (ODE) sampler. DDIM reinterprets the reverse diffusion process as a deterministic mapping, allowing high-fidelity sample generation with substantially fewer steps than stochastic Langevin-based approaches. Without loss of generality, we take conditional sampling with a single condition $\bm{c}$ as an example.  
At each sampling step $t$, given the noisy representation $\bm{x}_t$ and the estimated noise $\bm{\epsilon}_{\bm{\theta}}$ predicted by the denoising network, DDIM~\citep{DDIM} updates the posterior sampling from timestep $t$ to timestep $t-1$ with:
\begin{equation}
\label{DDIMODE}
\mathbf{x}_{t-1} = \sqrt{\bar{\alpha}_{t-1}} \left( \frac{\mathbf{x}_t - \sqrt{1 - \bar{\alpha}_{t}} \, \mathbf{\epsilon_\theta}(\mathbf{x}_t, t, \bm{c})}{\sqrt{\bar{\alpha}_{t}}} \right) + \sqrt{1 - \bar{\alpha}_{t-1}} \, \mathbf{\epsilon_\theta}(\mathbf{x}_t, t, \bm{c}).
\end{equation}
DDIM also presents a generalized form to control the stochasticity in generation process as follows:
\begin{equation}
\label{DDIMSDE}
\mathbf{x}_{t-1} = \sqrt{\bar{\alpha}_{t-1}} \left( \frac{\mathbf{x}_t - \sqrt{1 - \bar{\alpha}_{t}} \, \mathbf{\epsilon_\theta}(\mathbf{x}_t, t, \bm{c})}{\sqrt{\bar{\alpha}_{t}}} \right) + \sqrt{1 - \bar{\alpha}_{t-1} - \eta^2} \, \mathbf{\epsilon_\theta}(\mathbf{x}_t, t, \bm{c}) + \eta \bm{z},
\end{equation}
where $\bm{z}\sim \mathcal{N}(\bm{0},\bm{I})$ is isotropic Gaussian noise introducing stochasticity and $\eta \geq 0$ is a hyperparameter controlling stochasticity.
When $\eta= 0$, DDIM yields a fully deterministic trajectory, where Eq.~\eqref{DDIMSDE} naturally recovers Eq.~\eqref{DDIMODE}. This formulation allows DDIM to synthesize samples by progressively denoising the input through a deterministic mapping guided by the learned noise estimates at each timestep. By eliminating the need for stochastic perturbations during sampling, DDIM substantially accelerates inference while preserving the fidelity of generated data.

The use of DDIM provides multiple advantages for modeling cardiovascular signals. First, it substantially reduces the number of sampling steps, enabling rapid generation of PPG, ECG, and BP waveforms without compromising the fidelity of essential morphological features. Furthermore, the deterministic sampling path ensures that quasiperiodic structures (e.g., the QRS complex in ECG signals or the systolic and diastolic peaks in BP waveforms) are consistently preserved in generation, maintaining both temporal coherence and physiological plausibility. 

\subsection{Quantitative Results}
We present a comparison of sampling efficiency between the DDIM sampler~\citep{DDIM} and the original DDPM sampler~\citep{DDPM2020} in Table~\ref{tab:sampling_efficiency}. Across a variety of generation tasks, DDIM achieves synthesis quality comparable to DDPM while requiring significantly fewer sampling steps. Specifically, DDIM with only 6 sampling steps matches the performance of DDPM operating with 50 steps, resulting in nearly a 10-fold improvement in inference speed (reducing the generation time from over 2.5 seconds to less than 0.4 seconds). This acceleration is particularly critical for real-time monitoring of cardiovascular signals via timely generation of morphologically accurate waveforms.

\subsection{Case Study}
We conduct case studies to further demonstrate the sampling efficiency of UniCardio. In Fig.~\ref{fig:stepwise_diffusion}, we present a sampling trajectory of UniCardio on PPG imputation task, where missing values are generated conditioned on observed PPG segments. 
Starting from Gaussian noise sampled from $p(\bm{x}_T) \sim \mathcal{N}(\bm{0},\bm{I})$, UniCardio rapidly reconstructs the large-scale structures of the PPG waveform within the first 4 sampling steps, and progressively refines small-scale morphological details during the final steps approaching $t=0$. At each intermediate timestep, the denoising and refinement process is clearly visible, highlighting the capability of UniCardio to generate physiologically plausible waveforms with high efficiency in a small number of steps.

We further present additional generation examples of UniCardio across PPG imputation, ECG imputation, PPG-to-ECG translation, and PPG-to-BP translation (Fig.~\ref{fig:stepwise_task}).
Remarkably, UniCardio employs a single network trained in a unified manner to reconstruct diverse target signals from the same Gaussian noise prior, completing each generation within only 6 sampling steps. These results demonstrate the versatile functions of UniCardio to adaptively model a wide range of conditional distributions across different cardiovascular signals without task-specific modifications or re-training.

\section{Pseudo Code of Training and Sampling}\label{secE}

To capture the massive conditional distributions among PPG, ECG, and BP signals, UniCardio is pre-trained under a unified regime. The model progressively learns from tasks involving one, two, or three observed condition modalities through a phased curriculum. In each training phase, we dynamically sample generation targets and condition sets, ensuring balanced coverage of signal restoration and modality translation tasks. The training process is detailed in Algorithm~\ref{alg:unified-training}.

During inference, UniCardio employs a subset-based sampling strategy~\citep{DDIM} to accelerate signal generation. Given the selected target modality and the observed condition modalities, the model progressively refines an initial Gaussian noise through deterministic updates. The subset of time steps is selected by linearly spacing the desired number of steps across the diffusion range $[0, T]$. The sampling procedure follows the deterministic rule described in Eq.~\eqref{DDIMODE}, and is summarized in Algorithm~\ref{alg:unicardio-subset-ddim}.

\newpage
\begin{algorithm}[H]
\caption{Unified Training Process of UniCardio}
\label{alg:unified-training}
\KwIn{Total training epochs $e$, number of training phases $k = 4$, batch size $B$}
\KwOut{Trained UniCardio model $\bm{\epsilon}_\theta$}

Initialize model parameters $\bm{\theta}$ and learning rate $\eta \leftarrow 10^{-3}$\; \\

\For{phase $i = 1$ to $k$}{
    Define phase-specific sampling probabilities for the number of conditions\; \\
    
    Define phase-specific threshold $\mathcal{T}$ between 0 and 1\; \\

    \For{epoch $j = \frac{(i-1)e}{k}$ to $\frac{ie}{k}$}{
        Update learning rate $\eta$ according to the phase-specific schedule\; \\

        \For{each training iteration}{
            Sample one batch of clean PPG, ECG, and BP signals \; \\
            
            Construct the noisy version and the signals with missing values \; \\
            
            Sample the number of conditions for this training batch based on phase-specific sampling probabilities\; \\ 

            Sample from $\mathcal{U}(0,1)$ and compare with the threshold $\mathcal{T}$, determining the current batch to be restoration or translation task
            \; \\


            Determine the combination of condition and target modalities with the number of conditions and task type, and setup the task-specific attention mask\; \\

            

            Combine the sampled clean signals, noisy version, and the signals with missing values to create the condition $\bm{c}$ and target $\bm{x}_0$\; \\ 
            
            
            Sample timestep $t \sim \mathcal{U}(0, T]$\; \\

            Sample noise $\bm{\epsilon}_t \sim \mathcal{N}(\bm{0}, \bm{I})$\; \\
            
            Corrupt the target $\bm{x}_0$ using $\bm{\epsilon}_t$ to obtain noisy input $\bm{x}_t$\; \\

            Compute batch loss: \\
            \quad $\mathcal{L} = \frac{1}{B} \sum_{(\bm{x}_t, \bm{c}, t, \bm{\epsilon}_t)} \left\| \bm{\epsilon}_\theta(\bm{x}_t, \bm{c}, t) - \bm{\epsilon}_t \right\|^2$\; \\
            Update model parameters $\bm{\theta}$ using SGD with learning rate $\eta$\;
        }
    }
}
\Return{Trained UniCardio model $\bm{\epsilon}_\theta$}
\end{algorithm}

\begin{algorithm}[H]
\caption{Efficient Sampling Process of UniCardio}
\label{alg:unicardio-subset-ddim}
\KwIn{Trained UniCardio model $\bm{\epsilon}_\theta$, selected generation target modality, observed condition inputs $\bm{c}$, noise schedule $\{\bar{\alpha}_t\}_{t=1}^T$, a set of discrete timesteps $\{t\}$ linearly spaced from $T$ to $0$}
\KwOut{Generated target signal $\hat{\bm{x}}_0$}

Select the target modality to be generated, which can be PPG, ECG, or BP\; 

Assemble the condition set $\bm{c}$ by providing the observed modalities\; \\

Sample the initial noise $\bm{x}_T$ from the standard Gaussian distribution $\mathcal{N}(\bm{0}, \bm{I})$\; \\

\For{$t = T, T-1, \dots, 0$}{
    Predict the noise component: \\
    \quad $\hat{\bm{\epsilon}} = \bm{\epsilon}_\theta(\bm{x}_t, t, \bm{c})$\; \\

    If $t > 0$, update the next representation deterministically using Eq.~\eqref{DDIMODE}: \\
    \quad $\bm{x}_{t-1} = \sqrt{\bar{\alpha}_{t-1}} \left( \frac{\bm{x}_t - \sqrt{1 - \bar{\alpha}_{t}} \, \hat{\bm{\epsilon}}}{\sqrt{\bar{\alpha}_{t}}} \right) + \sqrt{1 - \bar{\alpha}_{t-1}} \, \hat{\bm{\epsilon}}$\;
}

\Return{Generated target signal $\hat{\bm{x}}_0$}
\end{algorithm}

\section{Additional Results}

\begin{figure}[ht]
    \centering
    \vspace{-0.1cm}
    \includegraphics[width=0.80\linewidth]{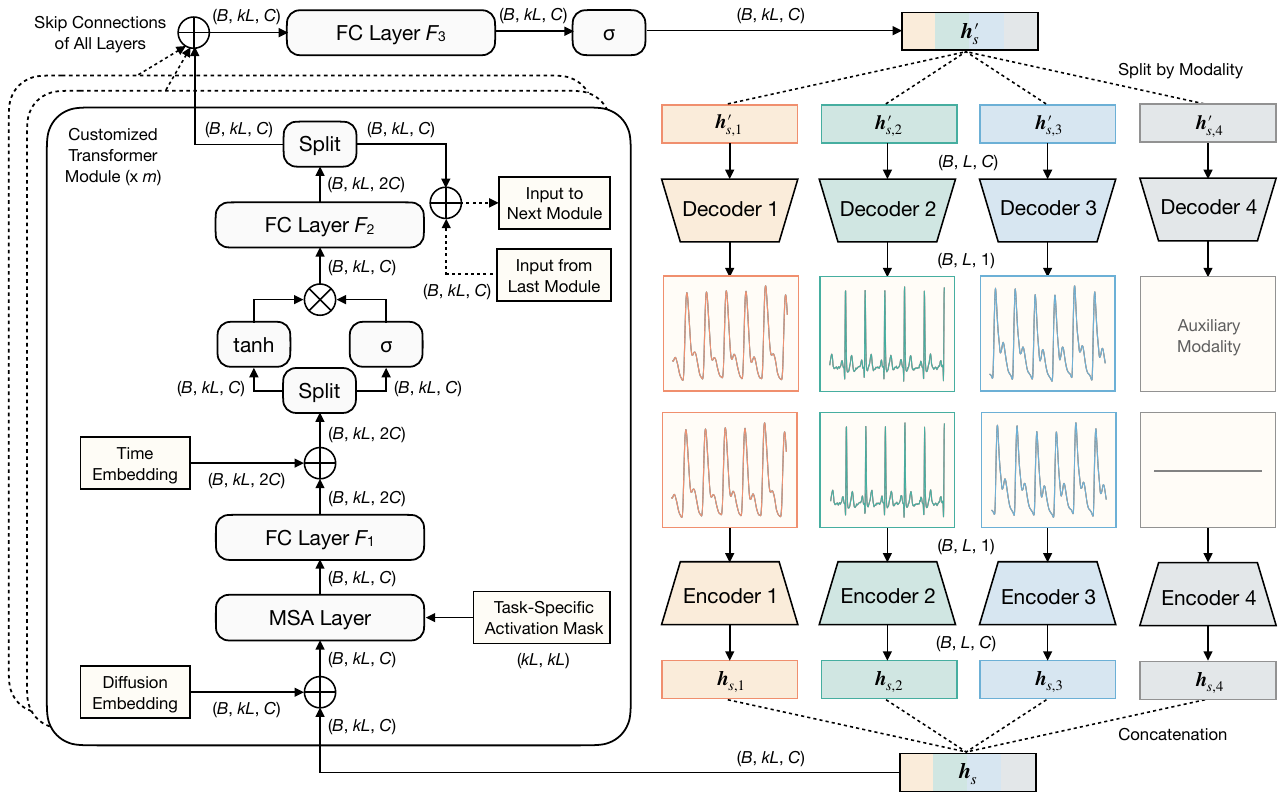}
    \caption{\textbf{Detailed model architecture of UniCardio.} Each modality-specific encoder consists of six consecutive 1D CNNs with various kernel sizes $\{1, 3, 5, 7, 9, 11\}$. The joint feature vector $\bm{h}_s$ is processed through five consecutive customized transformer modules with residual and skip connections, resulting in the final feature vector $\bm{h}_s'$. Each modality-specific decoder is implemented as a two-layer MLP with ReLU activation.
    } 
    \label{fig:model_architecture}
    \vspace{-0.1cm}
\end{figure}

\begin{table*}[ht]
    \centering
    \footnotesize 
    \caption{\textbf{Quantification results of phase-wise performance.} LRS, learning rate scheduling. TBC, training batch composition. TAM, task-specific attention mask. RMSE, the root mean squared error between generated signals and ground-truth signals. The quantification results are averaged by 256 independent trials. The error bars represent the standard error of the mean.}
    \smallskip
    \renewcommand\arraystretch{1.15}
     \addtolength{\tabcolsep}{-2pt}
    \resizebox{0.98\textwidth}{!}{ 
    \begin{tabular}{c|c|c|c|c|c}
    \hline
     Task & Method & Phase-1 RMSE ($\downarrow$)& Phase-2 RMSE ($\downarrow$)& Phase-3 RMSE ($\downarrow$)& Phase-4 RMSE ($\downarrow$)\\
     \hline
    \multirow{4}*{\tabincell{c}{PPG \\ Imputation}} 
     & UniCardio & 0.1160 $\pm 0.0066$ & 0.1144 $\pm 0.0064$ & 0.1180 $\pm 0.0065$& 0.1120 $\pm 0.0063$\\
     & UniCardio w/o LRS & 0.1461 $\pm 0.0063$& 0.1856 $\pm 0.0070$ & 0.3834 $\pm 0.0060$ & -- \\ 
     & UniCardio w/o TBC & 0.1093 $\pm 0.0062$ & 0.3791 $\pm 0.0056$ & 0.4125 $\pm 0.0037$ & -- \\
     & UniCardio w/o TAM & 1.9365 $\pm 0.0198$ & 60.7303 $\pm 0.7026$ & 52.8700 $\pm 0.3954$ & -- \\ 
    \hline 
    \multirow{4}*{\tabincell{c}{ECG \\ Imputation}} 
     & UniCardio & 0.1755 $\pm 0.0043$ & 0.1700 $\pm 0.0047$ & 0.1882 $\pm 0.0040$ & 0.1717 $\pm 0.0048$ \\
     & UniCardio w/o LRS & 0.2350 $\pm 0.0039$ & 0.2205 $\pm 0.0039$ & 0.2907 $\pm 0.0089$ & -- \\ 
     & UniCardio w/o TBC & 0.1783 $\pm 0.0045$ & 0.2916 $\pm 0.0034$ & 0.2944 $\pm 0.0034$ & -- \\
     & UniCardio w/o TAM & 3.0258 $\pm 0.0321$ & 9.3516 $\pm 0.2086$ & 34.6563 $\pm 0.6607$ & -- \\ 
    \hline 
     \multirow{4}*{\tabincell{c}{PPG-to-ECG \\ Translation}} 
     & UniCardio & 0.2945 $\pm 0.0063$ & 0.2873 $\pm 0.0071$ & 0.2871 $\pm 0.0068$ & 0.2759 $\pm 0.0067$ \\
     & UniCardio w/o LRS & 0.3581 $\pm 0.0097$ & 0.3397 $\pm 0.0054$ & 0.4916 $\pm 0.0089$ & -- \\ 
     & UniCardio w/o TBC & 0.2832 $\pm 0.0065$ & 0.3537 $\pm 0.0049$ & 0.3642 $\pm 0.0042$ & -- \\
     & UniCardio w/o TAM & 28.4177 $\pm 0.0547$ & 29.0761 $\pm 0.0410$ & 20.5246 $\pm 0.1508$ & -- \\ 
    \hline 
     \multirow{4}*{\tabincell{c}{PPG-to-BP \\ Translation}} 
     & UniCardio & 11.4886 $\pm 0.4465$ & 10.6768 $\pm 0.4491$ & 10.7759 $\pm 0.4433$ & 10.1675 $\pm 0.4153$ \\
     & UniCardio w/o LRS & 15.6685 $\pm 0.4075$ & 11.9641 $\pm 0.4083$ & 13.4415 $\pm 0.4052$ & -- \\ 
     & UniCardio w/o TBC & 10.7032 $\pm 0.4103$ & 15.0380 $\pm 0.4539$ & 20.0815 $\pm 0.4638$ & --  \\
     & UniCardio w/o TAM & 2601.45 $\pm 25.5475$ & 3046.11 $\pm 11.7325$ & 467.99 $\pm 8.7286$ & -- \\ 
     \hline
\end{tabular}
    \label{tab:forgetting}
}
\end{table*}

\begin{table*}[ht]
    \caption{
    \textcolor{darkgreen}{
    \textbf{Quantitative results of modality-restricted baselines.} Baseline-1 denotes single-modality models trained on one-condition imputation tasks. 
    Baseline-2 denotes dual-modality models trained on both one-condition translation tasks and two-condition imputation tasks. Baseline-3 denotes tri-modality models trained on both two-condition translation tasks and three-condition imputation tasks.
    All models use the same architecture, training data, and training paradigm to ensure fair comparison.
    The quantification results are averaged by 256 independent trials. The error bars represent the standard error of the mean.
    }
    }
    \centering
    \resizebox{0.95\textwidth}{!}{ 
    \begin{tabular}{c|c|c|c|c|c}
        \hline
        Task & Method & Input Signal & RMSE ($\downarrow$) & MAE ($\downarrow$) & KS-Test ($\downarrow$)  \\
        \hline
         \multirow{2}*{\tabincell{c}{PPG \\ Imputation}} 
         & Baseline-1 & PPG-I & $ 0.0924\pm0.0056 $ & $0.0671 \pm 0.0037 $ & $0.1032 \pm0.0033 $\\
         & UniCardio & PPG-I & $0.1146 \pm 0.0063 $ & $0.0797 \pm 0.0043 $ & $0.1084 \pm 0.0039 $ \\
        \cdashline{1-6}[2pt/2pt]
         \multirow{2}*{\tabincell{c}{ECG \\ Imputation}} 
         & Baseline-1 & ECG-I & $ 0.1319 \pm 0.0057 $ & $0.0632 \pm 0.0031$ & $0.1382 \pm 0.0048$\\
         & UniCardio  & ECG-I & $ 0.1756\pm0.0042 $ & $0.0750 \pm 0.0026 $ & $0.1486 \pm 0.0055 $ \\
        \cdashline{1-6}[2pt/2pt]
         \multirow{2}*{\tabincell{c}{BP \\ Imputation}} 
         & Baseline-1 & BP-I & $ 3.5269 \pm 0.2433 $ & $2.3243 \pm 0.1636 $ & $0.0982 \pm 0.0037 $\\
         & UniCardio  & BP-I & $4.3773 \pm 0.2896  $ & $ 2.8140 \pm 0.1960 $ & $0.1123 \pm 0.0041 $ \\
        \hline
        \multirow{2}*{\tabincell{c}{ECG \\ Imputation}} 
         & Baseline-2 & PPG, ECG-I & $0.0800 \pm 0.0039$ & $0.0428 \pm 0.0021$ & $0.1430 \pm 0.0040$\\
         & UniCardio & PPG, ECG-I & $0.0657 \pm 0.0039$ & $ 0.0344 \pm 0.0021$ & $0.1067 \pm 0.0035 $ \\
        \cdashline{1-6}[2pt/2pt]
        \multirow{2}*{\tabincell{c}{PPG \\ Imputation}} 
         & Baseline-2 & PPG-I, ECG & $0.0752 \pm 0.0039$ & $0.0583 \pm 0.0027$ & $0.1116 \pm 0.0039$\\
         & UniCardio & PPG-I, ECG & $0.0608 \pm 0.0031 $ & $ 0.0471 \pm 0.0021 $ & $ 0.1010 \pm 0.0034 $ \\
        \cdashline{1-6}[2pt/2pt]
        \multirow{2}*{\tabincell{c}{PPG-to-ECG \\ Translation}} 
         & Baseline-2 & PPG & $0.2587 \pm 0.0067$ & $0.1778 \pm 0.0071$ & $0.4085 \pm 0.0151$\\
         & UniCardio & PPG &  $0.2747 \pm 0.0067 $ & $ 0.1937 \pm 0.0070 $ & $0.4407 \pm 0.0154 $ \\
        \cdashline{1-6}[2pt/2pt]
        \multirow{2}*{\tabincell{c}{ECG-to-PPG \\ Translation}} 
         & Baseline-2 & ECG & $0.1539 \pm 0.0051$ & $0.1202 \pm 0.0040$ & $0.1406 \pm 0.0045$\\
         & UniCardio & ECG & $0.1802 \pm 0.0060 $ & $0.1390 \pm 0.0047 $ & $0.1491 \pm 0.0046 $ \\
          \hline
        \multirow{2}*{\tabincell{c}{BP \\ Imputation}} 
         & Baseline-2 & PPG, BP-I & $1.9208 \pm 0.1260$ & $1.4118 \pm 0.0688$ & $0.0939 \pm 0.0033$ \\ 
         & UniCardio & PPG, BP-I & $1.3995 \pm 0.1049$ & $ 1.0188 \pm 0.0555$ & $0.0766 \pm 0.0027$ \\ 
        \cdashline{1-6}[2pt/2pt]
        \multirow{2}*{\tabincell{c}{PPG \\ Imputation}} 
         & Baseline-2 & PPG-I, BP & $0.0610 \pm 0.0027$ & $0.0482 \pm 0.0020$ & $0.1111 \pm 0.0042$ \\  
         & UniCardio & PPG-I, BP & $0.0445 \pm 0.0016$ & $0.0349 \pm 0.0012$ & $0.0915 \pm 0.0033$ \\  
        \cdashline{1-6}[2pt/2pt]
        \multirow{2}*{\tabincell{c}{PPG-to-BP \\ Translation}} 
         & Baseline-2 & PPG & $9.6240 \pm 0.4190$ & $7.8722 \pm 0.8473$ & $0.2448 \pm 0.0092 $ \\  
         & UniCardio & ECG & $10.1538 \pm 0.4213 $ & $8.3721 \pm 0.3889 $ & $0.2668 \pm 0.0097 $ \\
        \cdashline{1-6}[2pt/2pt]
        \multirow{2}*{\tabincell{c}{BP-to-PPG \\ Translation}} 
         & Baseline-2 & PPG & $0.1511 \pm 0.0054$ & $0.1195 \pm 0.0042$ & $0.1311 \pm 0.0046$ \\  
         & UniCardio & PPG & $0.1471 \pm 0.0054$ & $0.1164 \pm 0.0043$ & $ 0.1361 \pm 0.0046$ \\  
         \hline
         \multirow{2}*{\tabincell{c}{ECG \\ Imputation}} 
         & Baseline-2 & ECG-I, BP & $0.0543 \pm 0.0033$ & $0.0333 \pm 0.0184$ & $0.1546 \pm 0.0044$ \\  
         & UniCardio & ECG-I, BP & $0.0400 \pm 0.0030$ & $0.0247 \pm 0.0018$ & $0.1014 \pm 0.0032$ \\  
         \cdashline{1-6}[2pt/2pt]
         \multirow{2}*{\tabincell{c}{BP \\ Imputation}} 
         & Baseline-2 & ECG, BP-I & $2.3571 \pm 0.1607$ & $1.7513 \pm 0.1055$ & $0.1010 \pm 0.0382$ \\  
         & UniCardio & ECG, BP-I & $1.6834 \pm 0.1056$ & $1.2637 \pm 0.0735$ & $0.0831 \pm 0.0030$ \\  
         \cdashline{1-6}[2pt/2pt]
        \multirow{2}*{\tabincell{c}{ECG-to-BP \\ Translation}} 
         & Baseline-2 & ECG & $7.0076 \pm 0.3556$ & $5.8537 \pm 0.3120$ & $0.2070 \pm 0.0089$ \\  
         & UniCardio & ECG & $8.0747 \pm 0.4052$ & $6.7337 \pm 0.3518$ & $0.2272 \pm 0.0092$ \\ 
         \cdashline{1-6}[2pt/2pt]
         \multirow{2}*{\tabincell{c}{BP-to-ECG \\ Translation}} 
         & Baseline-2 & ECG & $0.1554 \pm 0.0081$ & $0.1110 \pm 0.0073$ & $0.3050 \pm 0.0143$ \\  
         & UniCardio & ECG & $0.1717 \pm 0.0080$ & $0.1255 \pm 0.0075$ & $0.3433 \pm 0.0150$ \\  
        \hline
        
         \multirow{2}*{\tabincell{c}{PPG \\ Imputation}} 
         & Baseline-3 & PPG-I, ECG, BP & $0.0541 \pm 0.0026$ & $0.0426 \pm 0.0018$ & $0.1012 \pm 0.0033$ \\  
         & UniCardio & PPG-I, ECG, BP& $0.0444 \pm 0.0017$ & $0.0349 \pm 0.0013$ & $0.0911 \pm 0.0032$ \\
         \cdashline{1-6}[2pt/2pt]
         \multirow{2}*{\tabincell{c}{BP\&ECG-to-PPG \\ Translation}} 
         & Baseline-3 & ECG, BP & $0.1833 \pm 0.0064$ & $0.1458 \pm 0.0052$ & $0.1562 \pm 0.0046$ \\  
         & UniCardio & ECG, BP & $0.1324 \pm 0.0044$ & $0.1048 \pm 0.0036$ & $0.1299 \pm 0.0046$ \\ 
         \hline
         
         \multirow{2}*{\tabincell{c}{ECG \\ Imputation}} 
         & Baseline-3 & PPG, ECG-I, BP & $0.0598 \pm 0.0033$ & $0.0324 \pm 0.0018$ & $0.1174 \pm 0.0037$ \\  
         & UniCardio & PPG, ECG-I, BP & $0.0386 \pm 0.0028$ & $0.0241 \pm 0.0017$ & $0.1017 \pm 0.0033$ \\
         \cdashline{1-6}[2pt/2pt]
         \multirow{2}*{\tabincell{c}{PPG\&BP-to-ECG \\ Translation}} 
         & Baseline-3 & PPG, BP & $0.2887 \pm 0.0094$ & $0.2217 \pm 0.0099$ & $0.4622 \pm 0.0158$ \\  
         & UniCardio & PPG, BP & $0.1664 \pm 0.0076$ & $0.1201 \pm 0.0070$ & $0.3352 \pm 0.0147$ \\  
         \hline
         
         \multirow{2}*{\tabincell{c}{BP \\ Imputation}} 
         & Baseline-3 & PPG, ECG, BP-I & $1.4198 \pm 0.0709$ & $1.0637 \pm 0.0450$ & $0.0782 \pm 0.0026$ \\  
         & UniCardio & PPG, ECG, BP-I & $1.0723 \pm 0.0437$ & $0.8336 \pm 0.0334$ & $0.0703 \pm 0.0023$ \\
         \cdashline{1-6}[2pt/2pt]
        \multirow{2}*{\tabincell{c}{PPG\&ECG-to-BP \\ Translation}} 
         & Baseline-3 & PPG, ECG & $11.6832 \pm 0.5013$ & $10.2349 \pm 0.4833$ & $0.3107 \pm 0.0112$ \\  
         & UniCardio & PPG, ECG & $6.4716 \pm 0.3442$ & $5.4830 \pm 0.3082$ & $0.2013 \pm 0.0085$ \\  
        \hline  
    \end{tabular}
    }
    \label{tab:joint_model}
\end{table*}

\begin{table*}[th]
    \centering
\footnotesize 
    \caption{\textcolor{darkgreen}{\textbf{Quantification results of generated signals.} The evaluation metrics include the root mean squared error (RMSE), mean absolute error (MAE), signal-to-noise ratio (SNR), and Kolmogorov-Smirnov test (KS-Test), averaged by 256 independent trials.
    ``-N'' denotes the noised signals. ``-I'' denotes the intermittent signals.}}
    \resizebox{0.95\textwidth}{!}{ 
    \begin{tabular}{c|c|c|c|c|c}
    \hline
     Task & Method & Input Signal & RMSE ($\downarrow$) & SNR ($\uparrow$) & KS-Test ($\downarrow$)  \\
     \hline
    \multirow{5}*{\tabincell{c}{PPG\\ Denoising \\ (15 dB SNR)}} 
     & Wavelet Denoising~\cite{sendur2003bivariate}  &  PPG-N &0.0825  &18.0851  &0.0740  \\
     & UniCardio &  PPG-N & 0.0285 & 27.3389 & 0.0385 \\
     & UniCardio-M &  PPG-N, ECG & 0.0274 & 27.6628 & 0.0380 \\
     & UniCardio-M &  PPG-N, BP & 0.0257 & 28.2883 & 0.0374 \\
     & UniCardio-M &  PPG-N, ECG, BP & 0.0256 & 28.3034 & 0.0376 \\
    \hline 
    \multirow{5}*{\tabincell{c}{ECG\\ Denoising \\ (15 dB SNR)}} 
    & Wavelet Denoising~\cite{sendur2003bivariate} &  ECG-N &0.0936  &16.5280  &0.1267  \\
    & UniCardio  & ECG-N & 0.0289 & 26.6413 & 0.0812 \\
    & UniCardio-M & PPG, ECG-N & 0.0279 & 26.9689 & 0.0789 \\
    & UniCardio-M & ECG-N, BP & 0.0266 & 27.4222 & 0.0788 \\
    & UniCardio-M & PPG, ECG-N, BP & 0.0264  & 27.4700 & 0.0789 \\
    \hline
    \multirow{5}*{\tabincell{c}{BP\\ Denoising \\ (15 dB SNR)}} 
     & Wavelet Denoising~\cite{sendur2003bivariate} &  BP-N &3.3672  &28.8702  &0.0862  \\
     & UniCardio &  BP-N & 1.0573 & 38.9173 & 0.0428 \\
     & UniCardio-M &  PPG, BP-N & 0.9333 & 40.0531 & 0.0408 \\
     & UniCardio-M &  ECG, BP-N & 0.9560 & 39.8357 & 0.0409 \\
     & UniCardio-M &  PPG, ECG, BP-N & 0.8756 & 40.6339 & 0.0393 \\
     \hline
     \hline
    Task & Method & Input Signal & RMSE ($\downarrow$) & MAE ($\downarrow$) & KS-Test ($\downarrow$) \\
    \hline
    \multirow{6}*{\tabincell{c}{PPG \\ Imputation}} 
    & GP Algorithm~\cite{gerchberg1974super,papoulis2003new} & PPG-I & 0.7572  & 0.6304  & 0.3785  \\
    & UniCardio & PPG-I & 0.1146 & 0.0818 & 0.1084 \\
    & UniCardio-M & PPG-I, ECG & 0.0607 & 0.0465 & 0.0996 \\
    & UniCardio-M & PPG-I, BP & 0.0446 & 0.0349 & 0.0918 \\
    & UniCardio-M & PPG-I, ECG, BP & 0.0443 & 0.0347 & 0.0907 \\
    \hline
    \multirow{6}*{\tabincell{c}{ECG \\ Imputation}} 
    & GP Algorithm~\cite{gerchberg1974super,papoulis2003new} & ECG-I & 0.8664 & 0.2706  &0.6150  \\
    & UniCardio & ECG-I & 0.1722 & 0.0848 & 0.1822 \\ 
    & UniCardio-M & PPG, ECG-I & 0.0659 & 0.0345 & 0.1075 \\
    & UniCardio-M & ECG-I, BP & 0.0399 & 0.0246 & 0.1012 \\
    & UniCardio-M & PPG, ECG-I, BP & 0.0385 & 0.0241 & 0.1024 \\
    \hline
    \multirow{6}*{\tabincell{c}{BP \\ Imputation}} 
    & GP Algorithm~\cite{gerchberg1974super,papoulis2003new} & BP-I &87.5992 &79.0766  &0.9089  \\
    & UniCardio & BP-I & 4.3977 &2.8213 & 0.1148 \\
    & UniCardio-M & PPG, BP-I & 1.3994 & 1.0226 & 0.0770 \\
    & UniCardio-M & ECG, BP-I & 1.6877 & 1.2678 & 0.0834 \\
    & UniCardio-M & PPG, ECG, BP-I & 1.0827 & 0.8402 & 0.0697 \\
     \hline
\end{tabular}
\label{tab:classical_methods}
}
\end{table*}

\begin{table*}[t]
\centering
    \caption{\textbf{Comparison of sampling efficiency.} We evaluate the time required to generate each signal segment of 4 seconds with one-card A800 GPU. \textcolor{darkgreen}{The accompanying pre-processing pipeline adds only 0.0034 seconds per segment on a standard CPU of the same server.}
    NFE, the number of function evaluations for diffusion models. RMSE, the root mean squared error between generated signals and ground-truth signals.} 
	\smallskip
      \renewcommand\arraystretch{1.15}
     \addtolength{\tabcolsep}{-2pt}
	\resizebox{0.95\textwidth}{!}{ 
	\begin{tabular}{c|ccc|ccc|ccc|ccc}
	 \hline
        & \multicolumn{3}{c|}{PPG Imputation} & \multicolumn{3}{c|}{ECG Imputation} & \multicolumn{3}{c|}{PPG-to-ECG} & \multicolumn{3}{c}{PPG-to-BP} \\
        Sampler &NFE &RMSE ($\downarrow$) &Time ($\downarrow$) &NFE &RMSE ($\downarrow$)&Time ($\downarrow$)&NFE &RMSE ($\downarrow$)&Time ($\downarrow$)&NFE &RMSE ($\downarrow$)&Time ($\downarrow$)\\
        \hline
        DDPM &50 & 0.1119 & 2.81s &50 & 0.1732 & 2.74s &50 & 0.2753 & 2.96s &50 & 10.27 & 2.85s \\
        DDIM &6 & 0.1205 & 0.376s &6 & 0.1755 & 0.393s &6 & 0.2786 & 0.381s &6 & 10.26 & 0.398s \\
        DDIM &4 & 0.1203 & 0.288s &4 & 0.1747 & 0.278s &4 & 0.2802 & 0.267s &4 & 10.43 & 0.280s \\
       \hline
	\end{tabular}
	} 
	\label{tab:sampling_efficiency}
\end{table*}

\begin{table*}[ht]
    \centering
\footnotesize 
    \caption{\textbf{Task-specific attention mask.} We present the token ranges that are set to zero for different condition and target modalities.}
    \smallskip
    \renewcommand\arraystretch{1.15}
     \addtolength{\tabcolsep}{-2pt}
    \resizebox{0.90\textwidth}{!}{ 
    \begin{tabular}{c|c|c|c|c}
    \hline
     Modality & PPG & ECG & BP & AM \\
     \hline
     Condition & $M[--, 0:L]$ & $M[--, L:2L]$ & $M[--, 2L:3L]$ & $M[--, 3L:4L]$ \\
     Target & $M[0:L, --]$ & $M[L:2L, --]$ & $M[2L:3L, --]$ & $M[3L:4L, --]$ \\
    \hline 
\end{tabular}
    \label{tab:attention}
}
\end{table*}

\begin{figure}[t]
    \centering
    \vspace{-0.1cm}
    \includegraphics[width=0.98\linewidth]{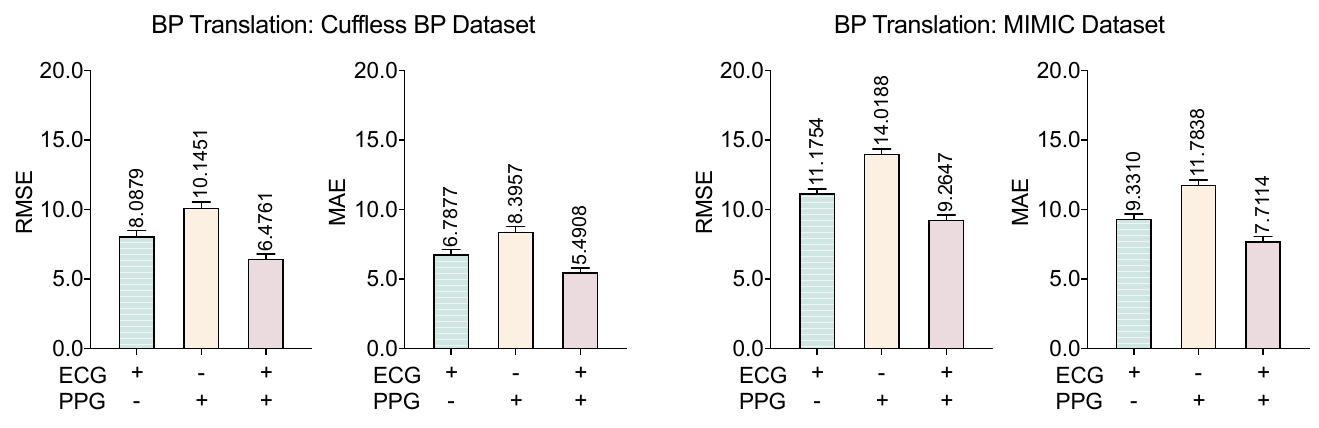}
    \caption{
    \textcolor{darkgreen}{
    \textbf{Performance of BP translation tasks.} 
    UniCardio is pretrained on the training set of Cuffless BP dataset. The generated signals are evaluated using unseen data of the Cuffless BP dataset~\cite{kachuee2016cuffless} (left) and MIMIC dataset~\cite{moody1996database} (right).
    The quantification results are averaged by 256 independent trials. The error bars represent the standard error of the mean.
    }
    } 
    \label{BP_Trans_Dataset}
    \vspace{-0.5cm}
\end{figure}

\begin{figure}
    \centering
    \includegraphics[width=1\linewidth]{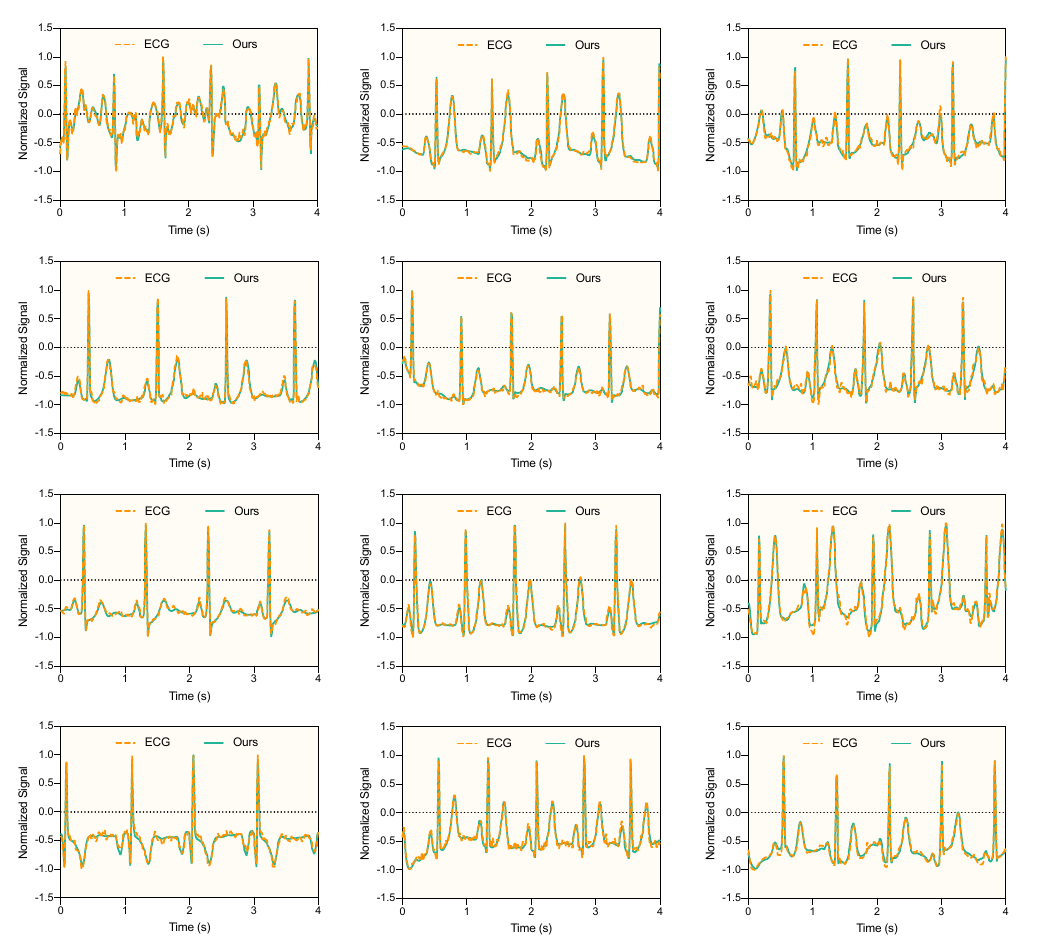}
    \caption{\textbf{Visualization of denoising results with ST change.} The ground-truth ECG signals are randomly selected from the PTBXL dataset~\cite{wagner2020ptb}. The generated signals are produced by UniCardio in a tuning-free manner.
 }
    \label{fig:denoising_st}
\end{figure}

\begin{figure}
    \centering
    \includegraphics[width=1\linewidth]{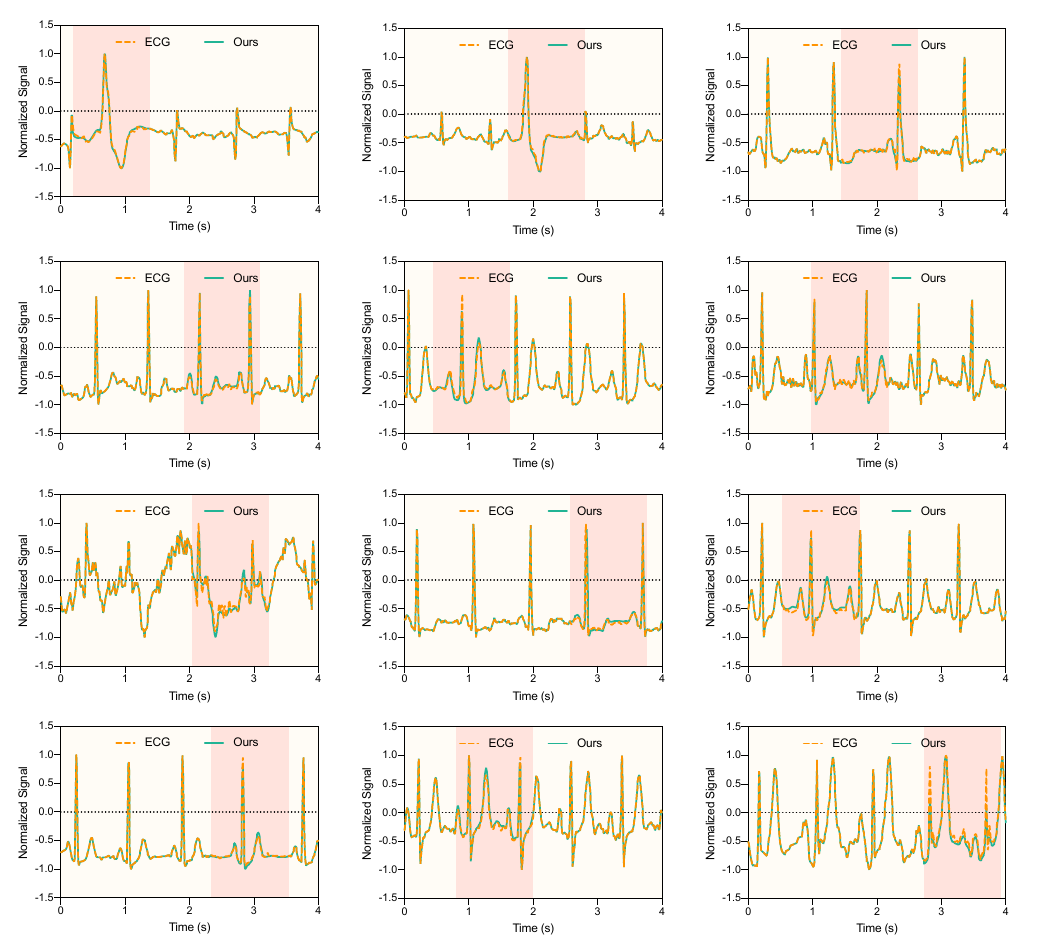}
    \caption{\textbf{Visualization of imputation results with ST change.} The ground-truth ECG signals are randomly selected from the PTBXL dataset~\cite{wagner2020ptb}. The generated signals are produced by UniCardio in a tuning-free manner. The masked regions denote the missing segments.
 }
    \label{fig:imputation_st}
\end{figure}

\begin{figure}
    \centering
    \includegraphics[width=1\linewidth]{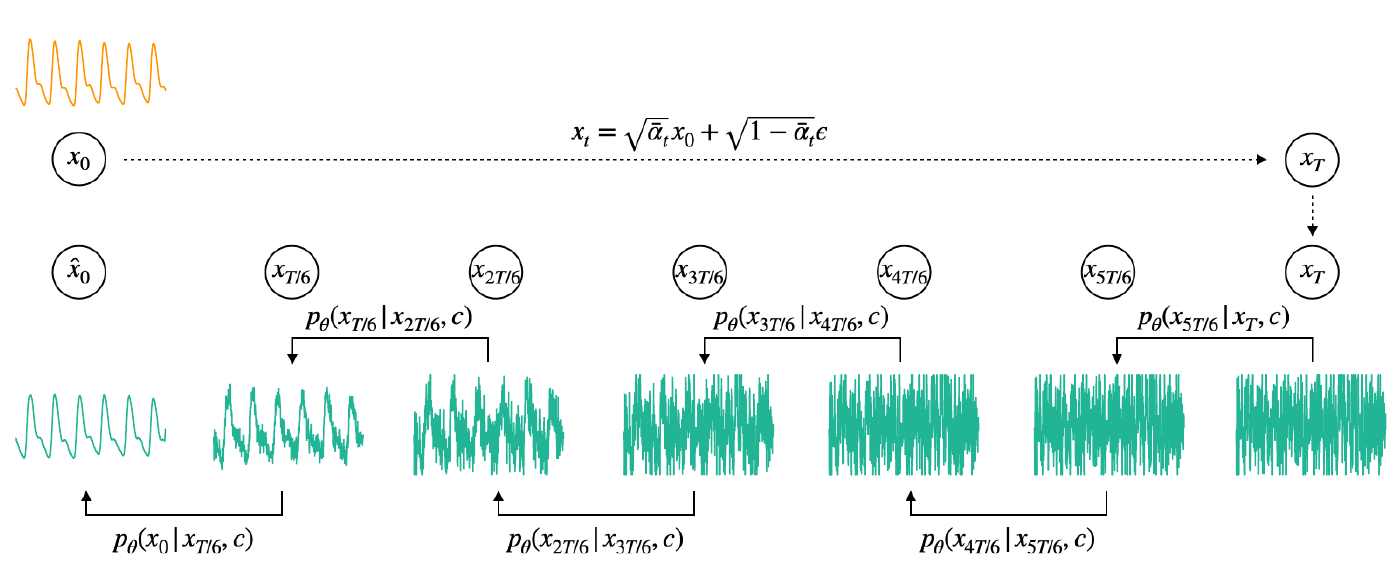}
    \caption{\textbf{Step-wise outputs of diffusion model.} We use 1/6 of all diffusion steps $T$ as the display interval. We set $T=50$ as the default implementation. More diffusion steps are closer to noise, while fewer diffusion steps are closer to clean signal. Here we take PPG imputation as the example of visualization.}
    \label{fig:stepwise_diffusion}
\end{figure}

\begin{figure}
    \centering
    \includegraphics[width=1\linewidth]{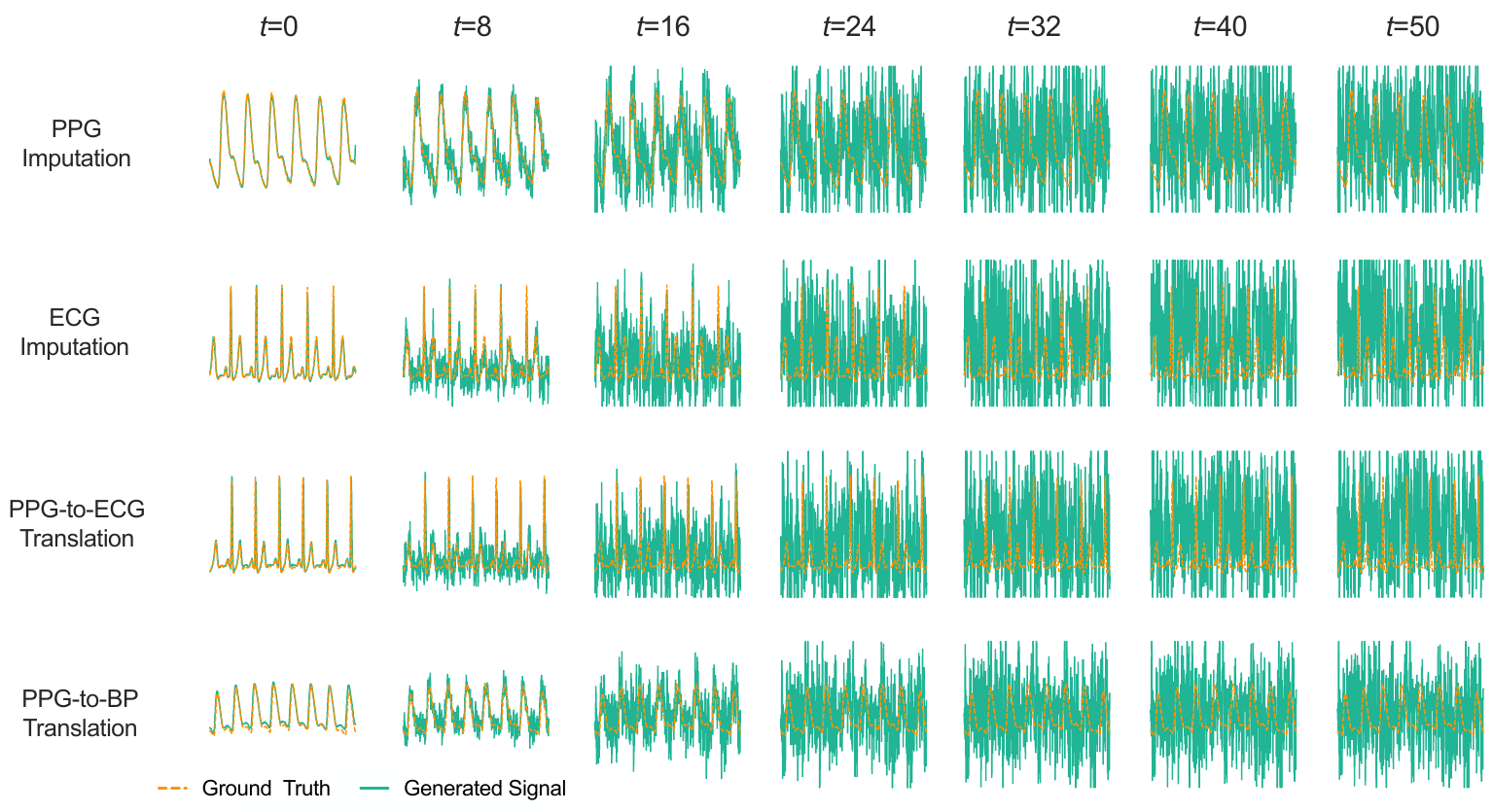}
    \caption{\textbf{Visualization of step-wise outputs.} Here we present visualization results of PPG imputation, ECG imputation, PPG-to-ECG translation, and PPG-to-BP translation tasks, with steps $t = \{0, 8, 16, 24, 32, 40, 50\}$.}
    \label{fig:stepwise_task}
\end{figure}

\end{document}